%% file: arxiv.tex
\documentclass{article}

\PassOptionsToPackage{numbers,sort&compress}{natbib}
\usepackage{PRIMEarxiv}

\usepackage{natbib}
\usepackage{amssymb}
\usepackage{amsthm}
\newtheorem{theorem}{Theorem}
\newtheorem{lemma}{Lemma}

\input{math_commands.tex}

\usepackage[utf8]{inputenc} 
\usepackage[T1]{fontenc}    
\usepackage{url}            
\usepackage{booktabs}       
\usepackage{amsfonts}       
\usepackage{nicefrac}       
\usepackage{microtype}      
\usepackage{xcolor}         
\usepackage{graphicx}
\graphicspath{{media/}{figures/}}
\usepackage{subcaption}
\usepackage[ruled,vlined]{algorithm2e}
\usepackage{bm}
\usepackage{comment}

\definecolor{citecolor}{HTML}{0071BC}
\definecolor{linkcolor}{HTML}{ED1C24}
\usepackage[pagebackref=false, breaklinks=true, letterpaper=true, colorlinks, bookmarks=false,  citecolor=citecolor, linkcolor=linkcolor, urlcolor=gray]{hyperref}

\newcommand{\papertitle}{Converge Then Diversify: Decoupling Convergence and Diversity in Multi-Objective Bayesian Optimisation}

\title{\papertitle}

\author{ Chao Jiang \\
	School of Computer Science\\
	University of Birmingham\\
	\texttt{cxj249@student.bham.ac.uk} \\
	\And
    Yueling Huang \\
	School of Computer Science\\
	University of Birmingham\\
	\texttt{yxh574@student.bham.ac.uk} \\
	\And
	Miqing Li\thanks{{Corresponding author: m.li.8@bham.ac.uk}} \\
	School of Computer Science\\
	University of Birmingham\\
	\texttt{m.li.8@bham.ac.uk} \\
}

\begin{document}

\maketitle

\begin{abstract}
Multi-objective Bayesian optimisation (MOBO) is a sample-efficient approach for optimising expensive black-box functions with multiple objectives. In MOBO, the goal is to adequately approximate the Pareto front; that is, to obtain a high-quality solution set with 1) good convergence (closeness to the Pareto front) and 2) good diversity (spread across the Pareto front). Existing MOBO methods typically aim to accomplish these two tasks simultaneously, i.e., driving the search towards the Pareto front while maintaining a diverse set of nondominated solutions, such that the solutions, ideally, can gradually approach the entire front. When sufficient search budgets are available, this approach is effective. However, considering both convergence and diversity throughout the search is not easy and requires careful design. Under very tight budgets, there may not be enough solutions generated to be able to simultaneously approach the entire Pareto front. To address this issue, this paper proposes a \textit{converge-then-diversify} (CTD) approach that decouples convergence and diversity into two stages. In the first stage, CTD focuses on convergence, aiming to quickly drive the search toward a single point on the Pareto front. In the second stage, CTD focuses on diversity, aiming to spread solutions across the front. We present two simple instantiations of CTD by using widely adopted acquisition functions in the area. Experimental results show that, across all 446 pairwise comparisons, CTD statistically outperforms state-of-the-art methods in 72.9\% of the cases, performs equivalently in 21.1\%, and is statistically worse in only 6.1\%, with the advantage being particularly evident in settings with very tight evaluation budgets or in high-dimensional problems.

\end{abstract}


\section{Introduction}


Multi-objective Bayesian optimisation (MOBO) is a sample-efficient approach for optimising expensive black-box functions with multiple objectives~\citep{garnett2023bayesian,wang2023recent,karl2023multi}. 
MOBO has been used in many areas, such as machine learning~\citep{jang2024model,wang2025trajectory}, physics~\citep{irshad2021expected,irshad2023multi}, and vehicle design~\citep{daulton2021parallel,anosri2023comparative}. 
In MOBO, the goal is to adequately approximate the Pareto front, that is, to obtain a high-quality solution set with both good convergence (closeness to the Pareto front) and good diversity (spread across the Pareto front).


To achieve this goal, various MOBO methods have been proposed. 
They can be loosely divided into scalarisation-based and Pareto-based methods. 
Scalarisation-based methods~\citep{knowles2006parego,paria2020flexible} convert a multi-objective problem into a number of single-objective subproblems by a set of weight vectors, and then use single-objective techniques to tackle these subproblems~\citep{chugh2020scalarizing}. 
In these methods, different weight vectors are used to guide the search towards different regions of the Pareto front.

Pareto-based methods consider Pareto dominance relations over objectives and directly solve a multi-objective problem without converting it into a number of single-objective subproblems. 
Representative examples include hypervolume-based and information-based methods. 
Hypervolume-based methods~\citep{emmerich2006single,daulton2020differentiable,daulton2021parallel} evaluate the hypervolume gain produced by each candidate, favouring points that either push the current front towards better objective values or enlarge its coverage over previously nondominated regions.
Information-based methods select points that reduce uncertainty about the Pareto front, favouring regions that are both promising and uncertain. 
These methods differ in the space over which information gain is measured: the input space of Pareto-optimal points~\citep{garrido2023parallel,garrido2019predictive,hernandez2016predictive}, the output space of the Pareto front~\citep{belakaria2019max,belakaria2021output,suzuki2020multi,ishikura2025paretofrontier}, or jointly over both~\citep{tu2022joint}. 
Apart from hypervolume-based and information-based methods, other Pareto-based methods select points using different criteria, such as maximum uncertainty~\citep{belakaria2020uncertainty} and maximin distance~\citep{renganathan2025qpots}, from a Pareto front obtained by optimising multiple acquisition functions (e.g., expected improvement~\citep{jones1998efficient} and Thompson sampling~\citep{thompson1933likelihood}).

All of the above methods, to obtain the Pareto front, consider convergence and diversity simultaneously throughout the search, aiming to approach multiple regions of the front. This strategy is effective when the evaluation budget is sufficient since maintaining a diverse set of solutions helps explore different areas. However, when the budget is very limited, balancing convergence and diversity becomes challenging. The number of evaluations is often insufficient to adequately approach the entire Pareto front. As a result, the search may yield a well-distributed set of nondominated solutions that are far from the true Pareto front.

\begin{figure}[t]
    \centering
    \includegraphics[width=0.7\linewidth]{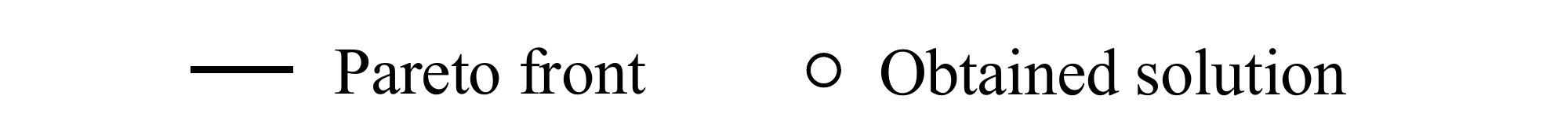}


    \begin{subfigure}[b]{0.48\linewidth}
        \centering
        \includegraphics[width=\linewidth]{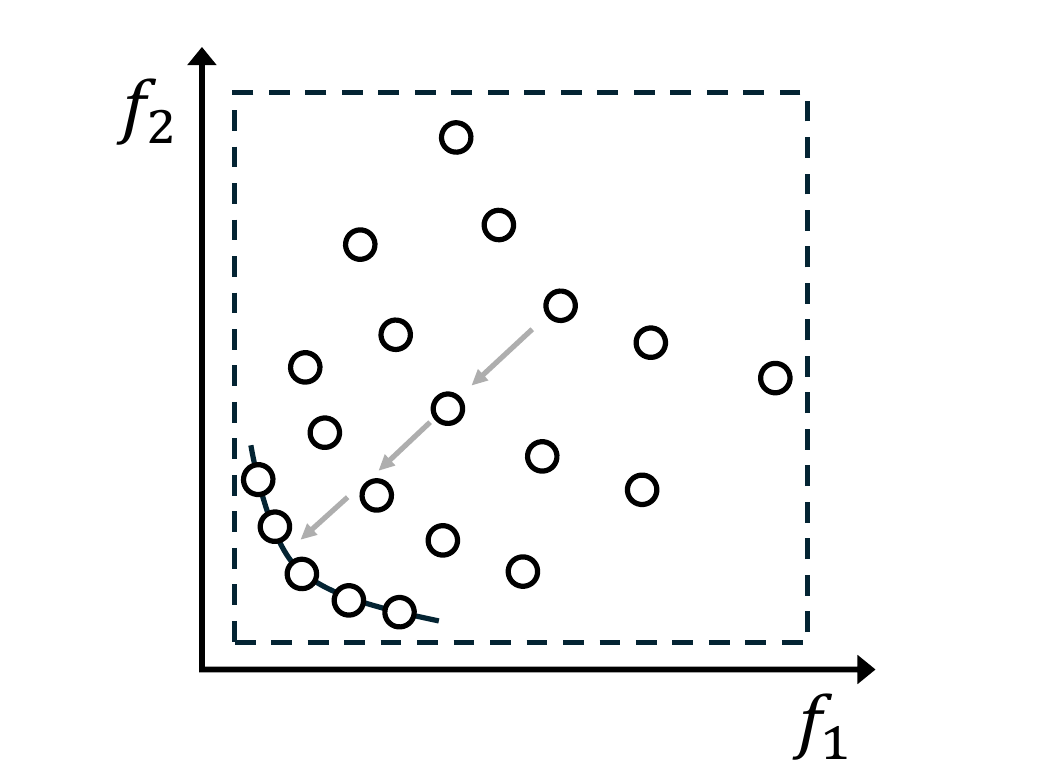}
        \caption{Conventional search in MOBO}
        \label{fig:other_method}
    \end{subfigure}
    \hfill
    \begin{subfigure}[b]{0.48\linewidth}
        \centering
        \includegraphics[width=\linewidth]{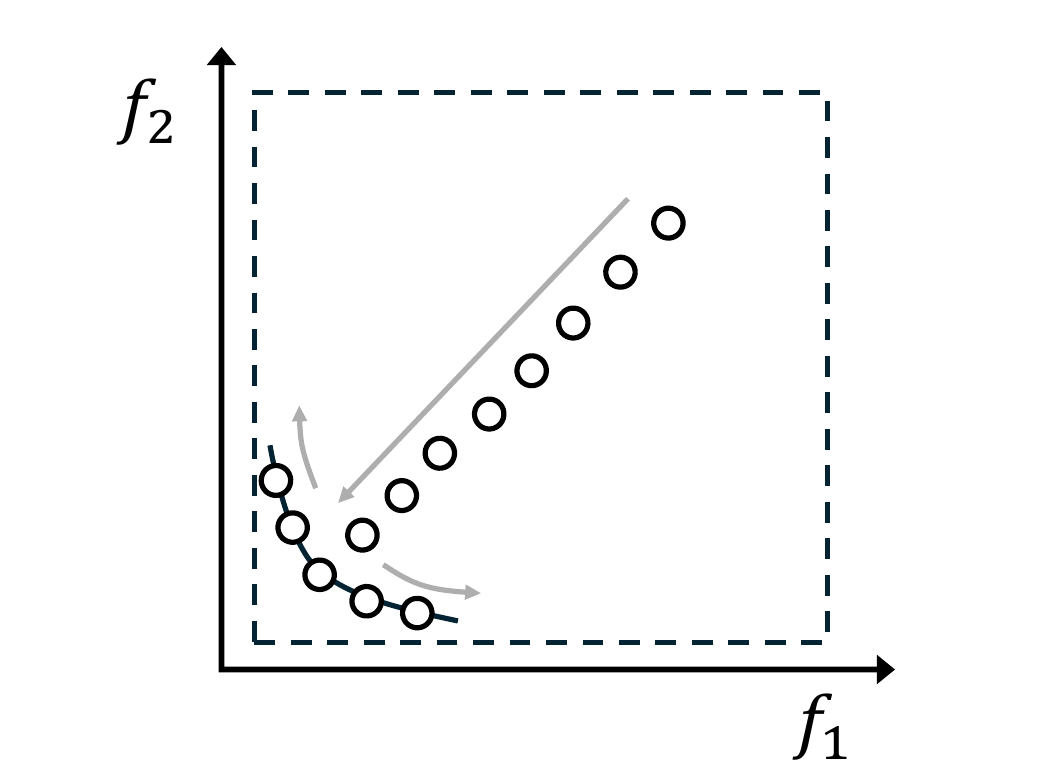}
        \caption{Proposed convergence-then-diversity search}
        \label{fig:proposed_method}
        
    \end{subfigure}

    \caption{Illustration of the search processes of the conventional MOBO approach and the proposed approach. (a) The conventional approach that moves towards the Pareto front while simultaneously spreading nondominated solutions. (b) The proposed \textit{convergence-then-diversity} strategy that first moves directly towards the Pareto front and then spreads along it.}
    \label{fig:illustration}
\end{figure}

To address this issue, this paper proposes an approach that decouples convergence from diversity, termed \textit{converge-then-diversify} (CTD). CTD operates in two stages. The first stage focuses on convergence, aiming to quickly drive the search toward a single point on the Pareto front. Once the solutions converge or are close to the front, the search switches to exploring along the front, promoting diversity. We hope that this design enables more efficient use of the evaluation budget while still obtaining a good approximation of the Pareto front.
Figure~\ref{fig:illustration} illustrates the proposed two-stage process in comparison with the conventional approach that optimises convergence and diversity simultaneously. 

The main contributions of this work are summarised as follows:

\begin{itemize}
    \item We propose a novel search framework for MOBO under very limited evaluation budgets, which decouples convergence and diversity into separate stages, enabling faster progress toward finding an approximation of the Pareto front.

    \item We show that the framework is general and flexible: it can be instantiated with different acquisition functions. In particular, we demonstrate its effectiveness when directly working with two widely adopted acquisition functions in the area.

    \item We validate the proposed framework through an extensive experimental study, comparing it with state-of-the-art methods under noiseless, noisy, and high-dimensional settings on both benchmark and real-world problems.

\end{itemize}

\section{Background}


\paragraph{Multi-Objective Optimisation.}  

Multi-objective optimisation refers to the process of optimising two or more objectives simultaneously, with the aim of obtaining a set of trade-off solutions rather than a single optimum. 
Without loss of generality, this paper considers the problem of minimising a vector-valued function: $\bm{f}(\bm{x}): \mathcal{X} \to \mathbb{R}^m$, where $\bm{x}\in \mathcal{X}$ ($\mathcal{X} \subset \mathbb{R}^{d}$) and $m$ is the number of objectives. 
In multi-objective optimisation, 
a solution $\bm x^*$ is said to dominate $\bm x'$, denoted by $\bm x^*\prec \bm x'$, if $\forall i \in\{1,\ldots,m\}$, $f_i(\bm x^*) \le f_i(\bm x')$ and $\exists j \in \{1,\ldots,m\}$, $f_j(\bm x^*) < f_j(\bm x')$. 
If a solution $\bm x^*\in\mathcal{X}$ is not dominated by any other solution, then $\bm x^*$ is said to be Pareto optimal. 
The collection of Pareto optimal solutions of a problem is called the Pareto set, and its mapping to the objective space is called Pareto front.  

To measure the quality of a Pareto front, hypervolume~\citep{Zitzler1999} is one of the most widely used metrics. Given a non-dominated set $\mathcal{P}_f$ and a reference point $\bm{r}$, the hypervolume is defined as $HV(\mathcal{P}_f,\bm{r})=\lambda_m\!\left(\bigcup_{\bm{p}\in\mathcal{P}_f}[\bm{r},\bm{p}]\right),$ where $[\bm{r},\bm{p}]$ denotes the hyperrectangle bounded by the reference point $\bm{r}$ and $\bm{p}$, and $\lambda_m(\cdot)$ denotes the $m$-dimensional Lebesgue measure.



\paragraph{Bayesian Optimisation (BO).} 

BO is a sample-efficient global optimisation method that builds a probabilistic surrogate, typically a Gaussian process (GP), and uses an acquisition function $\alpha: \mathcal{X} \to \mathbb{R}$ to decide which point to evaluate~\citep{garnett2023bayesian}. 
In this work, we model each objective with an independent Gaussian process $f_i \sim \mathcal{GP}(m_i(\bm x),k_i(\bm x,\bm x'))$, where $m_i: \mathcal{X} \to \mathbb{R}$ is the $i$th mean function, and $k_i(\cdot,\cdot): \mathcal{X} \times \mathcal{X} \to \mathbb{R}$ is the $i$th covariance function. 
Given $n$ observed points $\mathcal{D}^{(n)}= \{(\bm x^{(t)},\bm y^{(t)})\}_{t=1}^{n}$, where $\bm y^{(t)} = \bm{f}(\bm x^{(t)}) + \bm\zeta^{(t)}$ and the noise $\bm \zeta^{(t)} \sim \mathcal{N}(\bm 0, \text{diag} (\bm \sigma_\zeta^2))$, the posterior distribution of the $i$th objective at a new location $\bm x$ is Gaussian: 
$f_i(\bm x)\mid\mathcal{D}^{(n)} \sim \mathcal{N}(\mu_i(\bm x),\sigma_i^2(\bm x)),$ where $\mu_i(\bm x)$ and $\sigma_i^2(\bm x)$ are the posterior mean and variance at $\bm x$, respectively. 
Detailed expressions of the mean and variance are given in Appendix~\ref{appdx:sec:BO}.

Based on the posterior means and variances provided by the GP models, an acquisition function is optimised to determine the next solution to evaluate. 
Expected Improvement (EI)~\cite{mockus1978application,jones1998efficient,ament2023unexpected,jiang2025multi,jiang2025trading,jiang2026use} is one of the most prevalent acquisition functions. 
Considering a single-objective case, given the best observed value $y^{*}$, the improvement over $y^{*}$ is defined as $I(x)=\max (y^{*}-\mu(x),0)$. 
Then the EI can be expressed as: $\alpha_{EI}(x)=\mathbb{E}[I(x)]= \sigma(x)(\lambda\Phi(\lambda)+\phi(\lambda)),$ where $\lambda=\frac{y^{*}-\mu(x)}{\sigma(x)}$; $\Phi$ and $\phi$ are the standard normal cumulative distribution function and the probability density function, respectively. 
EI can be readily extended to multi-objective cases. Representative examples include scalarisation-based EI~\cite{knowles2006parego,zhang2009expensive} and hypervolume-based EI~\citep{emmerich2006single,daulton2020differentiable}.


\section{Related Work}

Over the past decades, various MOBO methods have been proposed~\citep{zuluaga2013active,shah2016pareto,daulton2022robust,tian2023autodex,yang2026multi,rashidi2024cylindrical,wang2023high,binois2020kalai,picheny2019bayesian,zhu2023sample,li2025expensive}. 
They can be loosely divided into scalarisation-based and Pareto-based methods. In scalarisation-based methods~\citep{knowles2006parego,paria2020flexible,zhang2024optimal}, a multi-objective problem is converted into a number of single-objective problems~\citep{chugh2020scalarizing,jiang2026we}. 
This allows acquisition functions from single-objective BO to be used for selecting the next point to evaluate. 
In these studies, different acquisition functions are employed, such as probability of improvement (PI)~\citep{kushner1964new} in~\citep{de2022mbore}, expected improvement (EI)~\citep{jones1998efficient} in~\citep{knowles2006parego,zhang2009expensive,namura2017expected,nogueirapires2025composite}, upper confidence bound (UCB)~\citep{lai1985asymptotically} in~\cite{paria2020flexible,zhang2020random,li2024constrained,lin2022pareto,li2025constrained,cheng2026parametric}, Thompson sampling (TS)~\citep{thompson1933likelihood} in~\cite{paria2020flexible,zhang2020random}, and knowledge gradient (KG)~\citep{frazier2008knowledge} in~\citep{buckingham2025knowledge}. 


Pareto-based methods consider Pareto dominance relations over objectives and directly solve a multi-objective problem without converting it into a number of single-objective subproblems. 
Representative examples include hypervolume-based and information-based methods. 
Hypervolume-based methods~\citep{ponweiser2008multiobjective,shang2020survey,bhatija2025multi,hung2025boformer,qing2023robust,zhang2025context} are prevalent since maximising the hypervolume value is equivalent to finding the entire Pareto front~\citep{zitzler2007hypervolume,li2019quality}. Many hypervolume-based methods are based on expected hypervolume improvement (EHVI), which extends expected improvement (EI) to the multi-objective setting~\citep{emmerich2006single}. 
EHVI has demonstrated strong performance across a range of settings~\citep{yang2016truncated,feliot2018ewhi,deng2025expected}, including sequential optimisation~\citep{emmerich2006single,rahat2022efficient}, batch optimisation \citep{yang2019multi2,wada2019bayesian,daulton2020differentiable,lu2024you}, constrained optimisation~\citep{abdolshah2018expected,de2022multi}, and noisy environments~\citep{daulton2021parallel}. 
Substantial efforts have also been devoted to improving its computational efficiency~\citep{couckuyt2014fast,hupkens2015faster,zhao2018fast,yang2019efficient,yang2019multi,mei2025pehvi}. 
Apart from the EHVI family, hypervolume-based methods have also been developed based on other acquisition functions, including
posterior mean~\citep{konakovic2020diversity}, probability of improvement (PI)~\citep{wang2022probability,couckuyt2014fast}, upper confidence bound (UCB)~\citep{roussel2021multiobjective,wang2026multiobjective}, Thompson sampling (TS)~\citep{bradford2018efficient,daulton2022multi}, and knowledge gradient (KG)~\citep{daulton2023hypervolume}.

Information-based methods leverage information theory to guide exploration toward regions likely to contribute to the Pareto front~\citep{qing2023pf,james2024multi,garrido2025information}. 
Representative approaches are predictive entropy search (PES), which focuses on the optimal inputs (i.e., the approximated Pareto set)~\citep{garrido2023parallel,garrido2019predictive,hernandez2016predictive}, max-value entropy search (MES), on the optimal outputs (i.e., the approximated Pareto front)~\citep{belakaria2019max,belakaria2021output,suzuki2020multi}, and joint entropy search (JES), on both of them~\citep{tu2022joint,fernandez2025joint}.


Apart from hypervolume-based and information-based methods, other Pareto-based methods select points according to other criteria~\citep{ahmadianshalchi2024pareto,park2023botied}. 
For example, some methods use maximum uncertainty~\citep{belakaria2020uncertainty}, while others use maximin distance~\citep{renganathan2025qpots} to select points from an approximate Pareto front obtained by optimising multiple acquisition functions for each objective, such as PI, EI, UCB, and TS.

All of the above methods consider convergence and diversity simultaneously across the search, i.e., guiding the search towards the Pareto front while maintaining a diverse set of nondominated solutions.



\begin{algorithm*}[t]
    \DontPrintSemicolon
    \caption{Converge-Then-Diversify (CTD) Framework}
    \label{alg:CTD}
    
    \KwIn {$\bm f$: Expensive black-box problem with $m$ objectives; $T$: Maximum number of iterations; \\
    \qquad \quad 
    $\alpha_{\text{converge}}$: Acquisition function in the convergence-focused stage; \\
    \qquad \quad $\alpha_{\text{diversify}}$: Acquisition function in the diversity-focused stage; \\
    \qquad \quad $\mathcal{D}^{(n_0)}:= \{(\bm x^{(t)},\bm{y}^{(t)})\}_{t=1}^{n_0} $: Initial observed solution set.
    

    }
    \nl \For{$t=n_0+1$ \KwTo $T$}{
        \nl $GPs \leftarrow$ Train $\mathcal{GP}s(\mathcal{D}^{(t-1)})$  \tcp*{Train $m$ Gaussian process models}
        \nl \eIf{convergence-focused stage}{
            \nl $\bm x^{(t)} \leftarrow \arg\max_{\bm x \in \mathcal{X}} \alpha_{\text{converge}}(\bm x, GPs)$ \tcp*{Convergence-focused search}

        }{
            \nl $\bm x^{(t)} \leftarrow \arg\max_{\bm x \in \mathcal{X}} \alpha_{\text{diversify}}(\bm x, GPs)$ \tcp*{Diversity-focused search}
        }
        
        \nl $\bm y^{(t)} \leftarrow \bm f(\bm x^{(t)}) + \bm \zeta^{(t)}$ \tcp*{Evaluate the selected solution}
        
        \nl $\mathcal{D}^{(t)} \leftarrow \mathcal{D}^{(t-1)} \cup \{(\bm x^{(t)},\bm y^{(t)})\}$ \tcp*{Augment the observed solution set}
    }


    \KwOut{$\mathcal{D}^{(T)}$: Observed solution set.}

\end{algorithm*}

\section{The Proposed Method}







Unlike existing MOBO methods, the proposed method decouples the tasks of converging into the Pareto front and diversifying along the Pareto front into two separate stages. In the first stage, it focuses on driving the search towards the Pareto front, and in the second stage, it focuses on spreading the solutions over the Pareto front. Algorithm~\ref{alg:CTD} gives the procedure of the proposed \textit{converge-then-diversify} (CTD) framework. As can be seen in the algorithm, CTD follows the basic procedure of MOBO. For each iteration, it trains $m$ Gaussian process models (Line 2), selects a solution for evaluation (Lines 3--5), and then evaluates the selected solution and augments the observed solution set (Lines 6 and 7). The key steps of CTD are the design in the convergence-focused stage (Line 4), the design in the diversity-focused stage (Line 5), and how to decide switching from convergence-focused search to diversity-focused search (Line 3). Next, we explain them individually in detail.

\subsection{Convergence-focused Search}

Designing the search strategy in this stage is straightforward. In principle, any acquisition function that can quickly guide the search towards one point of the Pareto front can be used. 
Here, we consider a well-known scalarisation-based acquisition function, EI based on the augmented Tchebycheff scalarisation~\citep{miettinen1999nonlinear}.
This scalarisation function has been frequently used in MOBO~\citep{balandat2020botorch,de2022mbore,ozaki2024multi}, including ParEGO~\citep{knowles2006parego} and PSL~\citep{lin2022pareto}. However, unlike these methods which consider different weight vectors to diversify the search directions for a well-distributed set of nondominated solutions, we consider a fixed weight vector, $(\frac{1}{m},\dots,\frac{1}{m}) \in \mathbb{R}^m$, where $m$ denotes the number of objectives. We hope this setting can enable the search quickly approach one point of the Pareto front, ideally a good trade-off solution among the objectives.

\subsection{Diversity-focused Search}

After the search converges to a region of the Pareto front, the focus shifts to diversifying the solutions. Note that diversification does not mean encouraging exploration in arbitrary directions - we do not want the search to move back towards regions that are dominated by solutions we found. Instead, the goal is to spread solutions along the Pareto front. 

Some existing acquisition functions can serve this purpose. The EI family is suitable since it seeks solutions that are expected to improve upon the current best solution (i.e., in our case, the converged solution already found). Integrated with a criterion that encourages diversity, EI would lead the search to expand from the converged solution. Acquisition functions of this type include hypervolume-based EI, because spreading solutions along the Pareto front always leads to improvements in hypervolume and the hypervolume value is maximised when the entire front is found. Similarly, scalarisation-based EI with diverse weight vectors can also serve the purpose, as it can lead the search to different regions of the Pareto front. In this paper, we adopt a classic hypervolume-based EI, EHVI~\citep{emmerich2006single,ament2023unexpected}. 
In the experimental study, we additionally consider a classic scalarisation-based EI (i.e., ParEGO) to further validate the proposed framework.

\subsection{Switching from Convergence to Diversity}


A critical issue in the proposed method is determining when to switch from convergence to diversity, i.e., when to stop the convergence-focused search. 
Stopping too early may result in solutions that are still far from the Pareto front, whereas stopping too late may lead to unnecessary exploration after the search has already converged. In this regard, the properties of EI come in handy, as a small maximum EI value indicates that further evaluations are unlikely to improve much~\citep{jones2001taxonomy}. Existing studies~\citep{jones2001taxonomy,nguyen2017regret} show that when EI is smaller than a small positive constant, the search should be stopped:

\begin{lemma}[Nguyen et al.~\citep{nguyen2017regret}]
\label{lem:ei-stopping}
The value of the EI acquisition function at the selected point should be positive for a valid optimisation, i.e.,
$\forall \bm x^{(t)} \in \mathcal{D}^{(t)}, \alpha_{EI}^{(t)}(\bm x^{(t)}) \geq \kappa > 0,$
where $\kappa$ is a small positive constant. 
If this condition is violated, the optimisation should be stopped.
\end{lemma}

In our context, this means the search has reached, or is very close to, a point of the Pareto front, since for a weight vector only a Pareto optimal solution can obtain the optimum of the augmented Tchebycheff scalarisation~\citep{steuer1986multiple}.
Specifically, according to~\citep{nguyen2017regret}, we set $\kappa=10^{-4}$ (i.e., the maximum EI value is smaller than $10^{-4}$) as the switching criterion.
In addition, it is worth mentioning that other stopping rules in BO~\citep{lorenz2015stopping,frazier2008knowledge,xie2025cost} (e.g., regret-based rules~\citep{makarova2022automatic,ishibashi2023stopping,wilson2024stopping}) may also be considered.

\vspace{-5pt}
\section{Experimental Design}\label{sec:exp_design}

\vspace{-5pt}
\paragraph{Compared Methods.} 

To evaluate the proposed CTD, we compare it with nine well-established methods: five Pareto-based methods, three scalarisation-based methods, and one baseline, Sobol~\citep{sobol1967distribution}. 
The five Pareto-based methods include three HV-based methods, i.e., EHVI~\citep{daulton2020differentiable,ament2023unexpected} (along with its noisy variant NEHVI~\citep{daulton2021parallel}), MORBO~\citep{daulton2022multi}, and MOBO-OSD~\citep{ngo2025moboosd}; one information-based method, JES~\citep{tu2022joint}; and one maximin-distance-based method, qPOTS~\citep{renganathan2025qpots}. 
The three scalarisation-based methods are ParEGO~\citep{knowles2006parego,ament2023unexpected} (along with its noisy variant NParEGO~\citep{daulton2021parallel}), TS-TCH~\citep{paria2020flexible}, and PSL~\citep{lin2022pareto}. 
For all EI-based methods, namely EHVI, NEHVI, ParEGO, NParEGO, and CTD, we use their log variants for numerical stability, as suggested in~\citep{ament2023unexpected}. 
Detailed implementations of our method and the peer methods can be found in Appendix~\ref{appdx:sec:implementation}. 

For all the methods, we sample $2(d + 1)$ initial points from a scrambled Sobol sequence, following the practice in~\citep{daulton2020differentiable,daulton2021parallel}. 
For methods using HV, i.e., EHVI, NEHVI, MORBO, PSL, MOBO-OSD, and CTD, we consider a setting frequently considered in the literature~\citep{daulton2020differentiable,tu2022joint}. That is, the reference point is
$\boldsymbol{r}
=
\boldsymbol{y}_{\text{nadir}}
+
0.1
\left|
\boldsymbol{y}_{\text{nadir}}
-
\boldsymbol{y}_{\text{ideal}}
\right|,$
where
$\boldsymbol{y}_{\text{nadir}}
=
(\max_{y_1\in\mathcal{D}^{(t)}_1} y_1,\ldots,\max_{y_m\in\mathcal{D}^{(t)}_m} y_m),$ 
$\boldsymbol{y}_{\text{ideal}}
=
(\min_{y_1\in\mathcal{D}^{(t)}_1} y_1,\ldots,\min_{y_m\in\mathcal{D}^{(t)}_m} y_m),$ and  $\mathcal{D}^{(t)}_i=\{y_i \mid \boldsymbol{y}\in\mathcal{D}^{(t)}\}$ denotes the set of observed values for the $i$th objective in the current observed dataset $\mathcal{D}^{(t)}$.


\begin{table*}[!ht]
\centering
\caption{The HV results (mean and standard deviation) of the ten methods on the 20 benchmark and real-world problems. 
The method with the best mean HV is highlighted in bold. The symbols ``$-$'', ``$\sim$'' and ``$+$'' indicate that our method CTD is statistically worse than, equivalent to, and better than the competitor, respectively.}
\label{tab:noiseless_100}
\resizebox{\textwidth}{!}{
\begin{tabular}{l | ll| ll| ll| ll| ll| ll|ll}
\toprule
\bfseries Method & \multicolumn{2}{c|}{\bfseries DTLZ1 ($m=2$)} & \multicolumn{2}{c|}{\bfseries DTLZ2 ($m=2$)} & \multicolumn{2}{c|}{\bfseries DTLZ3 ($m=2$)} & \multicolumn{2}{c|}{\bfseries DTLZ4 ($m=2$)} & \multicolumn{2}{c|}{\bfseries DTLZ5 ($m=2$)} & \multicolumn{2}{c|}{\bfseries DTLZ6 ($m=2$)} & \multicolumn{2}{c}{\bfseries DTLZ7 ($m=2$)} \\
 & \multicolumn{1}{c}{Mean} & \multicolumn{1}{c|}{Std} & \multicolumn{1}{c}{Mean} & \multicolumn{1}{c|}{Std} & \multicolumn{1}{c}{Mean} & \multicolumn{1}{c|}{Std} & \multicolumn{1}{c}{Mean} & \multicolumn{1}{c|}{Std} & \multicolumn{1}{c}{Mean} & \multicolumn{1}{c|}{Std} & \multicolumn{1}{c}{Mean} & \multicolumn{1}{c|}{Std} & \multicolumn{1}{c}{Mean} & \multicolumn{1}{c}{Std} \\ \midrule
\bfseries Sobol & 0.00e+00 & 0.00e+00 ($\sim$) & 1.23e-02 & 1.82e-02 ($+$) & 0.00e+00 & 0.00e+00 ($\sim$) & 1.34e-02 & 2.21e-02 ($+$) & 1.20e-02 & 1.33e-02 ($+$) & 0.00e+00 & 0.00e+00 ($+$) & 0.00e+00 & 0.00e+00 ($+$) \\
\bfseries ParEGO & 0.00e+00 & 0.00e+00 ($\sim$) & 2.97e-01 & 7.15e-02 ($+$) & 1.94e+02 & 5.61e+02 ($\sim$) & 2.84e-01 & 6.52e-02 ($+$) & 2.39e-01 & 1.06e-01 ($+$) & 2.67e-01 & 1.34e-01 ($\sim$) & 3.86e-01 & 2.38e-01 ($-$) \\
\bfseries TS-TCH & 0.00e+00 & 0.00e+00 ($\sim$) & 3.68e-02 & 3.99e-02 ($+$) & 0.00e+00 & 0.00e+00 ($\sim$) & 3.47e-02 & 3.43e-02 ($+$) & 4.32e-02 & 4.18e-02 ($+$) & 0.00e+00 & 0.00e+00 ($+$) & 0.00e+00 & 0.00e+00 ($+$) \\
\bfseries PSL & \textbf{7.92e+00} & \textbf{2.82e+01} ($\sim$) & 8.61e-02 & 5.44e-02 ($+$) & 2.66e+02 & 7.12e+02 ($\sim$) & 9.88e-02 & 3.64e-02 ($+$) & 8.61e-02 & 5.44e-02 ($+$) & 0.00e+00 & 0.00e+00 ($+$) & 4.64e-01 & 1.71e-01 ($-$) \\
\bfseries JES & 0.00e+00 & 0.00e+00 ($\sim$) & 2.43e-01 & 6.00e-02 ($+$) & 5.37e+02 & 6.34e+02 ($\sim$) & 2.65e-01 & 5.99e-02 ($+$) & 2.74e-01 & 4.40e-02 ($+$) & 2.22e-01 & 7.26e-02 ($+$) & 3.72e-01 & 1.86e-01 ($-$) \\
\bfseries EHVI & 0.00e+00 & 0.00e+00 ($\sim$) & 1.63e-01 & 1.15e-01 ($+$) & 5.54e+02 & 9.03e+02 ($\sim$) & 1.25e-01 & 9.14e-02 ($+$) & 1.32e-01 & 1.15e-01 ($+$) & 2.26e-01 & 3.17e-02 ($+$) & 1.69e-01 & 2.30e-02 ($\sim$) \\
\bfseries MORBO & 0.00e+00 & 0.00e+00 ($\sim$) & 4.07e-02 & 3.46e-02 ($+$) & 0.00e+00 & 0.00e+00 ($\sim$) & 3.91e-02 & 3.80e-02 ($+$) & 4.98e-02 & 3.98e-02 ($+$) & 0.00e+00 & 0.00e+00 ($+$) & 0.00e+00 & 0.00e+00 ($+$) \\
\bfseries MOBO-OSD & 0.00e+00 & 0.00e+00 ($\sim$) & 1.60e-01 & 6.90e-02 ($+$) & \textbf{8.62e+02} & \textbf{8.90e+02} ($\sim$) & 1.10e-01 & 1.06e-01 ($+$) & 1.54e-01 & 5.76e-02 ($+$) & 1.57e-01 & 8.31e-02 ($+$) & \textbf{7.19e-01} & \textbf{2.86e-02} ($-$) \\
\bfseries qPOTS & 0.00e+00 & 0.00e+00 ($\sim$) & 6.94e-03 & 1.19e-02 ($+$) & 0.00e+00 & 0.00e+00 ($\sim$) & 3.99e-03 & 9.05e-03 ($+$) & 4.09e-03 & 8.35e-03 ($+$) & 0.00e+00 & 0.00e+00 ($+$) & 0.00e+00 & 0.00e+00 ($+$) \\
\bfseries CTD (ours) & 6.68e+00 & 2.43e+01 & \textbf{3.46e-01} & \textbf{2.09e-02} & 4.17e+02 & 8.64e+02 & \textbf{3.52e-01} & \textbf{1.46e-02} & \textbf{3.49e-01} & \textbf{1.95e-02} & \textbf{3.19e-01} & \textbf{4.51e-02} & 1.71e-01 & 2.43e-02 \\
\bottomrule
\toprule
\bfseries Method & \multicolumn{2}{c|}{\bfseries DTLZ1 ($m=3$)} & \multicolumn{2}{c|}{\bfseries DTLZ2 ($m=3$)} & \multicolumn{2}{c|}{\bfseries DTLZ3 ($m=3$)} & \multicolumn{2}{c|}{\bfseries DTLZ4 ($m=3$)} & \multicolumn{2}{c|}{\bfseries DTLZ5 ($m=3$)} & \multicolumn{2}{c|}{\bfseries DTLZ6 ($m=3$)} & \multicolumn{2}{c}{\bfseries DTLZ7 ($m=3$)} \\
 & \multicolumn{1}{c}{Mean} & \multicolumn{1}{c|}{Std} & \multicolumn{1}{c}{Mean} & \multicolumn{1}{c|}{Std} & \multicolumn{1}{c}{Mean} & \multicolumn{1}{c|}{Std} & \multicolumn{1}{c}{Mean} & \multicolumn{1}{c|}{Std} & \multicolumn{1}{c}{Mean} & \multicolumn{1}{c|}{Std} & \multicolumn{1}{c}{Mean} & \multicolumn{1}{c|}{Std} & \multicolumn{1}{c}{Mean} & \multicolumn{1}{c}{Std} \\ \midrule
\bfseries Sobol & 1.19e+03 & 3.27e+03 ($\sim$) & 2.90e-02 & 2.10e-02 ($+$) & 0.00e+00 & 0.00e+00 ($+$) & 3.69e-02 & 2.42e-02 ($+$) & 1.12e-03 & 1.87e-03 ($+$) & 0.00e+00 & 0.00e+00 ($+$) & 0.00e+00 & 0.00e+00 ($+$) \\
\bfseries ParEGO & 4.31e+03 & 6.79e+03 ($\sim$) & 1.40e-01 & 1.20e-01 ($+$) & 9.97e+05 & 8.31e+05 ($\sim$) & 9.85e-02 & 1.12e-01 ($+$) & 6.85e-02 & 3.30e-02 ($+$) & 8.18e-02 & 3.32e-02 ($+$) & 4.87e-01 & 2.72e-01 ($-$) \\
\bfseries TS-TCH & 9.97e+02 & 4.40e+03 ($+$) & 3.23e-02 & 2.83e-02 ($+$) & 0.00e+00 & 0.00e+00 ($+$) & 2.56e-02 & 2.31e-02 ($+$) & 3.30e-03 & 6.49e-03 ($+$) & 0.00e+00 & 0.00e+00 ($+$) & 0.00e+00 & 0.00e+00 ($+$) \\
\bfseries PSL & 5.58e+03 & 6.89e+03 ($\sim$) & 6.66e-02 & 6.60e-03 ($+$) & 1.80e+06 & 6.38e+05 ($\sim$) & 5.71e-02 & 5.03e-02 ($+$) & 5.81e-02 & 2.42e-02 ($+$) & 0.00e+00 & 0.00e+00 ($+$) & 2.23e-01 & 1.25e-01 ($-$) \\
\bfseries JES & 8.38e+03 & 5.49e+03 ($-$) & 1.40e-01 & 1.04e-01 ($+$) & 1.77e+06 & 6.85e+05 ($\sim$) & 1.78e-01 & 1.23e-01 ($+$) & 5.04e-02 & 3.76e-02 ($+$) & 6.77e-02 & 2.40e-02 ($+$) & 3.93e-01 & 1.66e-01 ($-$) \\
\bfseries EHVI & 9.41e+03 & 5.68e+03 ($-$) & 4.56e-02 & 5.17e-02 ($+$) & 1.88e+06 & 7.78e+05 ($\sim$) & 4.70e-02 & 5.37e-02 ($+$) & 3.20e-02 & 3.18e-02 ($+$) & 6.67e-02 & 4.78e-03 ($+$) & 2.23e-01 & 1.73e-02 ($-$) \\
\bfseries MORBO & 9.50e+02 & 4.52e+03 ($+$) & 1.95e-02 & 2.32e-02 ($+$) & 0.00e+00 & 0.00e+00 ($+$) & 3.06e-02 & 3.52e-02 ($+$) & 1.22e-03 & 2.35e-03 ($+$) & 0.00e+00 & 0.00e+00 ($+$) & 0.00e+00 & 0.00e+00 ($+$) \\
\bfseries MOBO-OSD & \textbf{1.25e+04} & \textbf{4.79e+03} ($-$) & 1.77e-02 & 1.98e-02 ($+$) & \textbf{2.68e+06} & \textbf{1.01e+06} ($-$) & 1.52e-02 & 6.79e-02 ($+$) & 1.86e-02 & 2.78e-02 ($+$) & 2.68e-02 & 2.68e-02 ($+$) & \textbf{8.11e-01} & \textbf{7.21e-02} ($-$) \\
\bfseries qPOTS & 0.00e+00 & 0.00e+00 ($+$) & 0.00e+00 & 0.00e+00 ($+$) & 0.00e+00 & 0.00e+00 ($+$) & 0.00e+00 & 0.00e+00 ($+$) & 0.00e+00 & 0.00e+00 ($+$) & 0.00e+00 & 0.00e+00 ($+$) & 0.00e+00 & 0.00e+00 ($+$) \\
\bfseries CTD (ours) & 4.29e+03 & 5.30e+03 & \textbf{3.73e-01} & \textbf{7.95e-02} & 1.58e+06 & 8.91e+05 & \textbf{3.60e-01} & \textbf{9.01e-02} & \textbf{1.02e-01} & \textbf{1.30e-02} & \textbf{1.01e-01} & \textbf{1.73e-02} & 2.22e-01 & 1.90e-02 \\
\bottomrule
\toprule
\bfseries Method & \multicolumn{2}{c|}{\bfseries Four bar truss design} & \multicolumn{2}{c|}{\bfseries Pressure vessel design} & \multicolumn{2}{c|}{\bfseries Hatch cover design} & \multicolumn{2}{c|}{\bfseries Vehicle safety} & \multicolumn{2}{c|}{\bfseries Car side impact} & \multicolumn{2}{c|}{\bfseries LPA} & \multicolumn{2}{c}{\bfseries Summary} \\
 & \multicolumn{1}{c}{Mean} & \multicolumn{1}{c|}{Std} & \multicolumn{1}{c}{Mean} & \multicolumn{1}{c|}{Std} & \multicolumn{1}{c}{Mean} & \multicolumn{1}{c|}{Std} & \multicolumn{1}{c}{Mean} & \multicolumn{1}{c|}{Std} & \multicolumn{1}{c}{Mean} & \multicolumn{1}{c|}{Std} & \multicolumn{1}{c}{Mean} & \multicolumn{1}{c|}{Std} & \multicolumn{2}{c}{$-$/$\sim$/$+$} \\ \midrule
\bfseries Sobol & 4.45e+01 & 6.61e-01 ($+$) & 3.90e+09 & 1.87e+09 ($+$) & 2.01e+04 & 7.43e+02 ($+$) & 1.48e+01 & 9.38e-01 ($+$) & 2.25e+02 & 5.17e+00 ($+$) & 2.53e+07 & 2.65e+06 ($+$) & \multicolumn{2}{c}{0/3/17} \\
\bfseries ParEGO & 5.19e+01 & 1.34e+00 ($+$) & 1.10e+10 & 5.81e+07 ($\sim$) & 2.18e+04 & 3.04e+01 ($+$) & 2.62e+01 & 5.52e-01 ($\sim$) & 2.96e+02 & 1.01e+02 ($+$) & 4.12e+07 & 3.08e+06 ($+$) & \multicolumn{2}{c}{4/5/11} \\
\bfseries TS-TCH & 5.07e+01 & 1.32e+00 ($+$) & 1.03e+10 & 2.79e+08 ($+$) & 2.16e+04 & 8.35e+01 ($+$) & 2.26e+01 & 3.76e-01 ($+$) & 2.54e+02 & 8.64e+01 ($+$) & 3.75e+07 & 3.11e+06 ($+$) & \multicolumn{2}{c}{0/2/18} \\
\bfseries PSL & 5.39e+01 & 3.87e-02 ($+$) & 1.09e+10 & 1.72e+08 ($\sim$) & 2.17e+04 & 1.42e+01 ($+$) & 2.37e+01 & 1.50e+00 ($+$) & 2.77e+02 & 1.09e+01 ($+$) & \textbf{4.81e+07} & \textbf{2.13e+06} ($\sim$) & \multicolumn{2}{c}{2/6/12} \\
\bfseries JES & 5.15e+01 & 1.62e+00 ($+$) & 1.08e+10 & 1.39e+08 ($+$) & 2.17e+04 & 4.07e+01 ($+$) & \textbf{2.66e+01} & \textbf{4.17e-01} ($-$) & \textbf{3.27e+02} & \textbf{9.44e+00} ($-$) & 4.03e+07 & 4.05e+06 ($+$) & \multicolumn{2}{c}{5/3/12} \\
\bfseries EHVI & 5.40e+01 & 9.20e-02 ($+$) & 1.09e+10 & 1.51e+08 ($\sim$) & \textbf{2.18e+04} & \textbf{1.83e+01} ($\sim$) & 2.54e+01 & 8.40e-01 ($+$) & 2.79e+02 & 9.52e+01 ($\sim$) & 4.75e+07 & 3.02e+06 ($\sim$) & \multicolumn{2}{c}{2/8/10} \\
\bfseries MORBO & 5.06e+01 & 1.23e+00 ($+$) & 1.03e+10 & 3.62e+08 ($+$) & 2.16e+04 & 9.12e+01 ($+$) & 2.24e+01 & 4.56e-01 ($+$) & 2.83e+02 & 5.65e+00 ($+$) & 3.73e+07 & 2.77e+06 ($+$) & \multicolumn{2}{c}{0/2/18} \\
\bfseries MOBO-OSD & 5.20e+01 & 8.56e-01 ($+$) & \textbf{1.10e+10} & \textbf{2.77e+07} ($\sim$) & 2.13e+04 & 3.62e+02 ($+$) & 2.35e+01 & 1.14e+00 ($+$) & 2.78e+02 & 9.90e+00 ($+$) & 3.65e+07 & 3.96e+06 ($+$) & \multicolumn{2}{c}{4/3/13} \\
\bfseries qPOTS & 3.97e+01 & 1.31e+00 ($+$) & 1.08e+10 & 1.93e+08 ($+$) & 1.85e+04 & 4.03e+03 ($+$) & 1.30e+01 & 2.37e+00 ($+$) & 2.43e+02 & 1.06e+01 ($+$) & 3.85e+07 & 2.16e+06 ($+$) & \multicolumn{2}{c}{0/2/18} \\
\bfseries CTD (ours) & \textbf{5.41e+01} & \textbf{2.82e-02} & 1.09e+10 & 1.79e+08 & 2.18e+04 & 2.05e+01 & 2.60e+01 & 6.38e-01 & 3.05e+02 & 1.55e+01 & 4.62e+07 & 3.71e+06 & \multicolumn{2}{c}{--/--/--} \\
\bottomrule
\end{tabular}
}
\end{table*}

\vspace{-5pt}
\paragraph{Benchmarks and Real-World Problems.} For benchmark problems, we choose the most widely used scalable functions, DTLZ1--DTLZ7~\citep{deb2005scalable}. 
Each problem is considered with 2 and 3 objectives, following the practice in~\citep{deb2013evolutionary,jain2013evolutionary,li2014evolutionary}.
We also consider six well-studied real-world problems~\citep{tanabe2020easy,irshad2023multi}, i.e., four bar truss design~\citep{cheng1999generalized}, pressure vessel design~\citep{kannan1994augmented}, hatch cover design~\citep{amir1989nonlinear}, vehicle safety design~\citep{liao2008multiobjective}, car side impact design~\citep{jain2013evolutionary}, and laser plasma acceleration (LPA)~\citep{irshad2021expected,irshad2023multi}. 
These problems are widely used in the MOBO literature~\citep{bradford2018efficient,daulton2020differentiable,daulton2021parallel,lin2022pareto,ament2023unexpected,renganathan2025qpots,yang2026multi}. 
The details of the problem formulations are given in Appendix~\ref{appendix:subsec:problems}. 
We test the above problems under both noiseless and noisy cases. 
To make their noise cases, we include additive zero-mean Gaussian noise with a standard deviation set to approximately 10\% of the objective ranges, as suggested in~\cite{daulton2021parallel,tu2022joint}.



\vspace{-5pt}
\paragraph{Statistical Validation.} 
To enable statistical comparisons, each optimisation was repeated 30 times. 
We use the Wilcoxon rank-sum test~\cite{wilcoxon1992individual} at a significance level of 0.05 and Holm-Bonferroni correction~\cite{holm1979simple} to see if our method differs significantly from each peer method. 

\paragraph{Performance Indicator.} 

HV~\cite{Zitzler1999} is used to compare all methods. 
For the evaluation, we use the commonly used setting in quality evaluation of multi-objective optimisation \citep{li2019quality}. 
That is, we collect all solutions obtained by all methods over 30 independent runs and set the reference point $\boldsymbol{r}
=
\boldsymbol{y}_{\text{nadir}}
+
0.1
\left|
\boldsymbol{y}_{\text{nadir}}
-
\boldsymbol{y}_{\text{ideal}}
\right|$. 

\section{Experimental Results}

\subsection{How Does CTD Perform Compared with  Well-Established Methods?}

\paragraph{Noiseless Cases.}

We begin our evaluation in the noiseless setting. 
We set the dimensionality to $d=m+4$ for DTLZ1, $d=m+9$ for DTLZ2--DTLZ6, and $d=m+19$ for DTLZ7, according to the original paper~\cite{deb2005scalable}. 
For all the methods, we consider a very limited budget, allowing a maximum of 100 evaluations, which is not an uncommon setting in the area (e.g.,~\citep{daulton2020differentiable,ip2025user}). 

Table~\ref{tab:noiseless_100} shows the HV results (mean and standard deviation) of our method and the nine peer methods on the 20 benchmark and real-world problems. 
As shown in the table, our method demonstrates a clear performance advantage, performing statistically better, equivalently, and worse in 129, 34, and 17 out of the 180 pairwise comparisons, respectively. 
Despite its strong overall performance, CTD performs relatively poorly on DTLZ7, which has a disconnected Pareto front. This is mainly because EHVI is used during the diversity-focused search, as EHVI itself struggles to explore the entire Pareto front of DTLZ7~\citep{chugh2022mono}. 
That said, CTD can, in principle, handle different Pareto-front structures because it does not restrict the search space during the diversity-focused stage. Evidence of this can be found in the pressure-vessel design problem, which also has a disconnected Pareto front~\citep{tanabe2020easy}. On this problem, where EHVI performs well, CTD is among the best-performing methods.

\begin{figure}[!ht]
    \centering
    \includegraphics[width=0.5\linewidth]{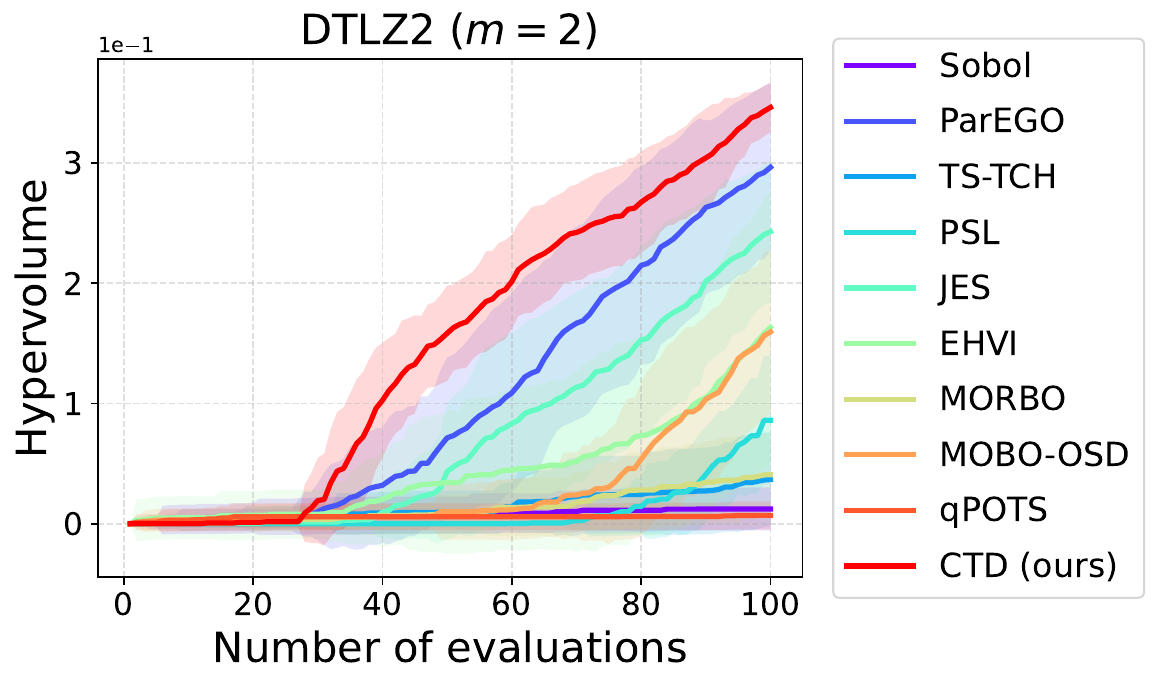}
    \caption{Mean hypervolume trajectories of the ten methods on the representative DTLZ2 problem ($m=2$, $d=11$) over 30 independent runs. The shaded regions indicate one standard deviation around the mean.}
    \label{fig:hv_traj}
\end{figure}

To help understand their search behaviour, Figure~\ref{fig:hv_traj} shows the hypervolume trajectories with respect to the number of evaluations of the ten methods on the problem DTLZ2 with $m=2$. 
As shown in the figure, since around 30 evaluations (notably, the first 24 evaluations are initial samples), CTD exhibits a clear advantage, obtaining substantially higher hypervolume values than its competitors.

\begin{figure*}[!ht]
    \centering

    \begin{subfigure}{\textwidth}
        \centering
        \includegraphics[width=\textwidth]{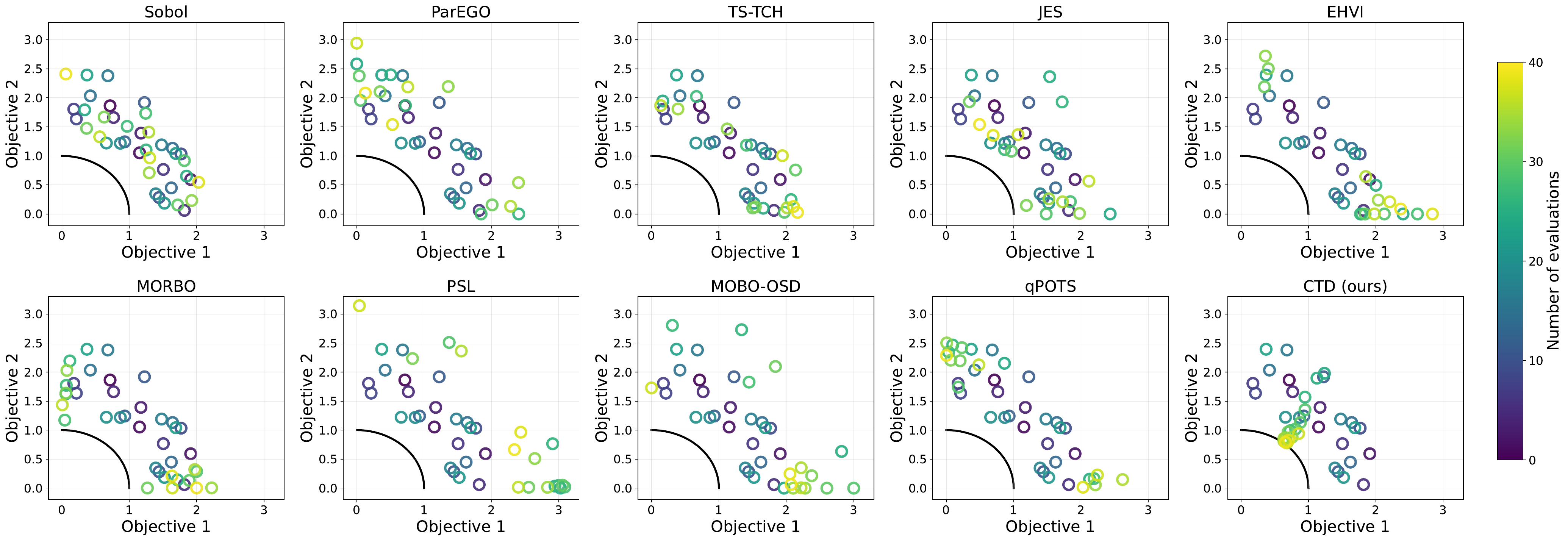}
        \caption{Solutions obtained by the proposed CTD and the nine peer methods under 40 evaluations.}
        \label{fig:dtlz2-budget40}
    \end{subfigure}

    \vspace{0.5em}

    \begin{subfigure}{\textwidth}
        \centering
        \includegraphics[width=\textwidth]{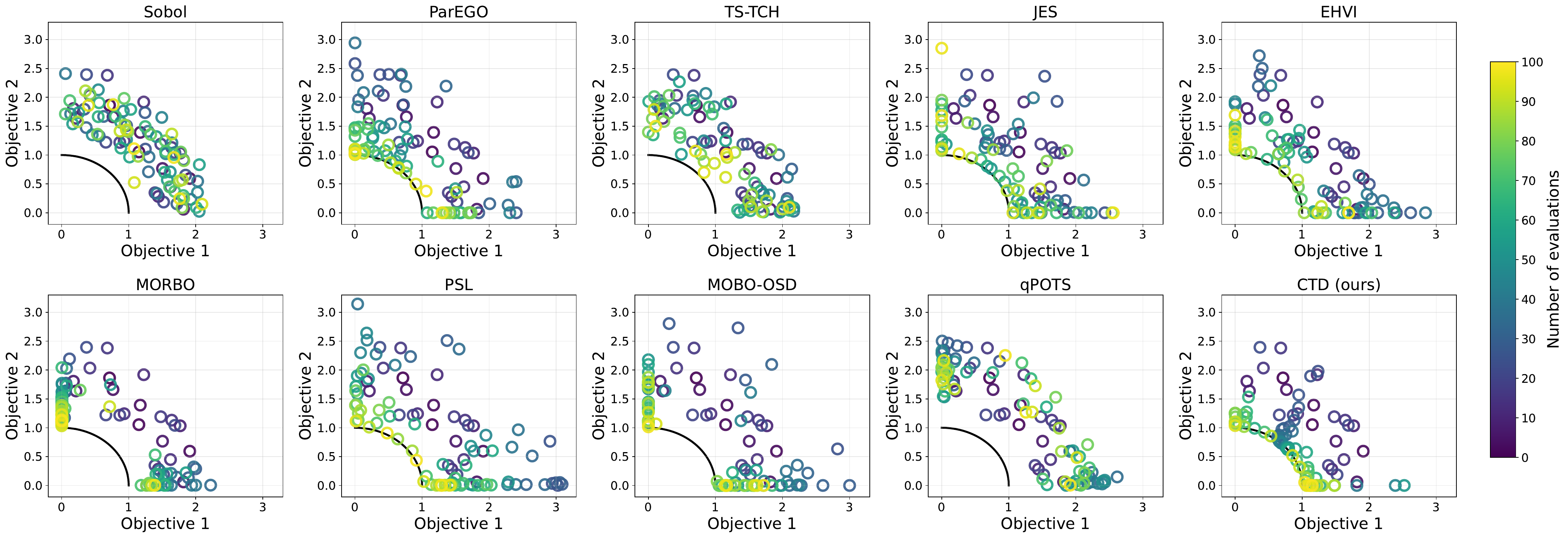}
        \caption{Solutions obtained by the proposed CTD and the nine peer methods under 100 evaluations.}
        \label{fig:dtlz2-budget100}
    \end{subfigure}

    \caption{Solutions obtained by the proposed CTD and the nine peer methods on DTLZ2 ($m=2$, $d=11$) under budgets of 40 and 100 function evaluations. All methods start from the same $2(d+1)$ initial points generated by a scrambled Sobol sequence. The black line denotes the Pareto front, and brighter colours indicate solutions obtained at later evaluations.}
    \label{fig:dtlz2-pf-comparison}
\end{figure*}

To visually understand the results, Figure~\ref{fig:dtlz2-pf-comparison} shows the solutions obtained by each method on DTLZ2 ($m=2$, $d=11$), where for facilitating comparison, all methods start from the same $2(d+1)$ initial points generated by a scrambled Sobol sequence. 
As shown in the plot (a), the convergence-focused search enables CTD to approach the Pareto front under a very limited budget, while the other methods remain distant from the Pareto front. 
Furthermore, after approaching the Pareto front, CTD is able to spread solutions along the front and achieve good diversity (plot (b) in Figure~\ref{fig:dtlz2-pf-comparison}).

\begin{table*}[!ht]\large
\centering
\caption{The HV results (mean and standard deviation) of the eight methods on the 20 benchmark and real-world problems under noisy settings. 
The method with the best mean HV is highlighted in bold. The symbols ``$-$'', ``$\sim$'' and ``$+$'' indicate that the proposed CTD is statistically worse than, equivalent to, and better than the competitor, respectively.}
\label{tab:noisy_100}
\resizebox{\textwidth}{!}{
\begin{tabular}{l | ll| ll| ll| ll| ll| ll|ll}
\toprule
\bfseries Method & \multicolumn{2}{c|}{\bfseries DTLZ1 ($m=2$)} & \multicolumn{2}{c|}{\bfseries DTLZ2 ($m=2$)} & \multicolumn{2}{c|}{\bfseries DTLZ3 ($m=2$)} & \multicolumn{2}{c|}{\bfseries DTLZ4 ($m=2$)} & \multicolumn{2}{c|}{\bfseries DTLZ5 ($m=2$)} & \multicolumn{2}{c|}{\bfseries DTLZ6 ($m=2$)} & \multicolumn{2}{c}{\bfseries DTLZ7 ($m=2$)} \\
 & \multicolumn{1}{c}{Mean} & \multicolumn{1}{c|}{Std} & \multicolumn{1}{c}{Mean} & \multicolumn{1}{c|}{Std} & \multicolumn{1}{c}{Mean} & \multicolumn{1}{c|}{Std} & \multicolumn{1}{c}{Mean} & \multicolumn{1}{c|}{Std} & \multicolumn{1}{c}{Mean} & \multicolumn{1}{c|}{Std} & \multicolumn{1}{c}{Mean} & \multicolumn{1}{c|}{Std} & \multicolumn{1}{c}{Mean} & \multicolumn{1}{c}{Std} \\ \midrule
\bfseries Sobol & 0.00e+00 & 0.00e+00 ($\sim$) & 1.54e-02 & 1.39e-02 ($+$) & 0.00e+00 & 0.00e+00 ($\sim$) & 1.37e-02 & 1.66e-02 ($+$) & 1.17e-02 & 1.61e-02 ($+$) & 0.00e+00 & 0.00e+00 ($+$) & 0.00e+00 & 0.00e+00 ($\sim$) \\
\bfseries NParEGO & 0.00e+00 & 0.00e+00 ($\sim$) & 6.94e-02 & 4.58e-02 ($\sim$) & 3.98e+02 & 4.96e+02 ($\sim$) & 5.68e-02 & 6.14e-02 ($\sim$) & \textbf{7.16e-02} & \textbf{6.93e-02} ($\sim$) & 1.36e-01 & 9.29e-02 ($\sim$) & 1.39e-02 & 7.63e-02 ($\sim$) \\
\bfseries TS-TCH & 0.00e+00 & 0.00e+00 ($\sim$) & 2.76e-03 & 4.60e-03 ($+$) & 0.00e+00 & 0.00e+00 ($\sim$) & 2.92e-03 & 8.28e-03 ($+$) & 5.82e-03 & 1.47e-02 ($+$) & 0.00e+00 & 0.00e+00 ($+$) & 0.00e+00 & 0.00e+00 ($\sim$) \\
\bfseries JES & 0.00e+00 & 0.00e+00 ($\sim$) & 5.95e-02 & 5.53e-02 ($\sim$) & 3.24e+02 & 7.44e+02 ($\sim$) & 3.93e-02 & 4.22e-02 ($\sim$) & 4.20e-02 & 4.51e-02 ($\sim$) & 1.12e-01 & 9.06e-02 ($\sim$) & 0.00e+00 & 0.00e+00 ($\sim$) \\
\bfseries NEHVI & 0.00e+00 & 0.00e+00 ($\sim$) & 2.38e-02 & 4.35e-02 ($+$) & \textbf{4.40e+02} & \textbf{6.41e+02} ($\sim$) & 2.82e-02 & 4.75e-02 ($+$) & 3.20e-02 & 5.52e-02 ($+$) & \textbf{1.71e-01} & \textbf{6.73e-02} ($\sim$) & \textbf{1.50e-02} & \textbf{7.13e-02} ($\sim$) \\
\bfseries MORBO & 0.00e+00 & 0.00e+00 ($\sim$) & 5.55e-04 & 1.62e-03 ($+$) & 0.00e+00 & 0.00e+00 ($\sim$) & 2.07e-03 & 7.19e-03 ($+$) & 3.47e-03 & 8.53e-03 ($+$) & 0.00e+00 & 0.00e+00 ($+$) & 0.00e+00 & 0.00e+00 ($\sim$) \\
\bfseries qPOTS & 0.00e+00 & 0.00e+00 ($\sim$) & 4.22e-03 & 8.03e-03 ($+$) & 0.00e+00 & 0.00e+00 ($\sim$) & 5.04e-03 & 1.39e-02 ($+$) & 5.77e-03 & 1.68e-02 ($+$) & 0.00e+00 & 0.00e+00 ($+$) & 0.00e+00 & 0.00e+00 ($\sim$) \\
\bfseries CTD (ours) & \textbf{8.72e-03} & \textbf{4.78e-02} & \textbf{7.59e-02} & \textbf{5.95e-02} & 3.69e+02 & 6.54e+02 & \textbf{5.68e-02} & \textbf{4.74e-02} & 6.70e-02 & 4.69e-02 & 1.33e-01 & 7.48e-02 & 1.48e-02 & 5.13e-02\\
\bottomrule
\toprule
\bfseries Method & \multicolumn{2}{c|}{\bfseries DTLZ1 ($m=3$)} & \multicolumn{2}{c|}{\bfseries DTLZ2 ($m=3$)} & \multicolumn{2}{c|}{\bfseries DTLZ3 ($m=3$)} & \multicolumn{2}{c|}{\bfseries DTLZ4 ($m=3$)} & \multicolumn{2}{c|}{\bfseries DTLZ5 ($m=3$)} & \multicolumn{2}{c|}{\bfseries DTLZ6 ($m=3$)} & \multicolumn{2}{c}{\bfseries DTLZ7 ($m=3$)} \\
 & \multicolumn{1}{c}{Mean} & \multicolumn{1}{c|}{Std} & \multicolumn{1}{c}{Mean} & \multicolumn{1}{c|}{Std} & \multicolumn{1}{c}{Mean} & \multicolumn{1}{c|}{Std} & \multicolumn{1}{c}{Mean} & \multicolumn{1}{c|}{Std} & \multicolumn{1}{c}{Mean} & \multicolumn{1}{c|}{Std} & \multicolumn{1}{c}{Mean} & \multicolumn{1}{c|}{Std} & \multicolumn{1}{c}{Mean} & \multicolumn{1}{c}{Std} \\ \midrule
\bfseries Sobol & 3.08e+02 & 1.35e+03 ($+$) & 3.16e-02 & 1.96e-02 ($+$) & 0.00e+00 & 0.00e+00 ($+$) & 3.56e-02 & 2.71e-02 ($+$) & 1.48e-03 & 3.22e-03 ($+$) & 0.00e+00 & 0.00e+00 ($+$) & 0.00e+00 & 0.00e+00 ($\sim$) \\
\bfseries NParEGO & 5.77e+03 & 5.09e+03 ($\sim$) & 6.26e-02 & 6.14e-02 ($+$) & 1.05e+06 & 7.39e+05 ($\sim$) & 5.32e-02 & 4.45e-02 ($+$) & 7.85e-03 & 1.08e-02 ($+$) & 2.53e-02 & 3.25e-02 ($+$) & 0.00e+00 & 0.00e+00 ($\sim$) \\
\bfseries TS-TCH & 1.18e+03 & 2.40e+03 ($+$) & 8.98e-03 & 1.37e-02 ($+$) & 0.00e+00 & 0.00e+00 ($+$) & 1.29e-02 & 1.40e-02 ($+$) & 2.89e-04 & 1.04e-03 ($+$) & 0.00e+00 & 0.00e+00 ($+$) & 0.00e+00 & 0.00e+00 ($\sim$) \\
\bfseries JES & 6.80e+03 & 5.15e+03 ($\sim$) & 5.61e-02 & 4.62e-02 ($+$) & 7.39e+05 & 6.05e+05 ($+$) & 5.40e-02 & 4.67e-02 ($+$) & 4.93e-03 & 9.71e-03 ($+$) & 1.27e-02 & 2.50e-02 ($+$) & 0.00e+00 & 0.00e+00 ($\sim$) \\
\bfseries NEHVI & \textbf{9.89e+03} & \textbf{5.04e+03} ($\sim$) & 1.61e-02 & 2.48e-02 ($+$) & \textbf{1.87e+06} & \textbf{6.86e+05} ($-$) & 1.96e-02 & 3.03e-02 ($+$) & 9.73e-04 & 2.28e-03 ($+$) & 3.72e-02 & 3.40e-02 ($\sim$) & 0.00e+00 & 0.00e+00 ($\sim$) \\
\bfseries MORBO & 7.29e+02 & 3.38e+03 ($+$) & 7.14e-03 & 9.57e-03 ($+$) & 0.00e+00 & 0.00e+00 ($+$) & 1.67e-02 & 2.77e-02 ($+$) & 2.68e-04 & 7.75e-04 ($+$) & 0.00e+00 & 0.00e+00 ($+$) & 0.00e+00 & 0.00e+00 ($\sim$) \\
\bfseries qPOTS & 4.85e+02 & 1.53e+03 ($+$) & 1.26e-02 & 1.53e-02 ($+$) & 0.00e+00 & 0.00e+00 ($+$) & 8.72e-03 & 1.11e-02 ($+$) & 2.56e-04 & 7.47e-04 ($+$) & 0.00e+00 & 0.00e+00 ($+$) & 0.00e+00 & 0.00e+00 ($\sim$) \\
\bfseries CTD (ours) & 8.33e+03 & 6.55e+03 & \textbf{9.15e-02} & \textbf{4.19e-02} & 1.43e+06 & 6.62e+05 & \textbf{8.17e-02} & \textbf{4.46e-02} & \textbf{1.47e-02} & \textbf{1.20e-02} & \textbf{5.16e-02} & \textbf{3.09e-02} & \textbf{0.00e+00} & \textbf{0.00e+00} \\
\bottomrule
\toprule
\bfseries Method & \multicolumn{2}{c|}{\bfseries Four bar truss design} & \multicolumn{2}{c|}{\bfseries Pressure vessel design} & \multicolumn{2}{c|}{\bfseries Hatch cover design} & \multicolumn{2}{c|}{\bfseries Vehicle safety} & \multicolumn{2}{c|}{\bfseries Car side impact} & \multicolumn{2}{c|}{\bfseries LPA} & \multicolumn{2}{c}{\bfseries Summary} \\
 & \multicolumn{1}{c}{Mean} & \multicolumn{1}{c|}{Std} & \multicolumn{1}{c}{Mean} & \multicolumn{1}{c|}{Std} & \multicolumn{1}{c}{Mean} & \multicolumn{1}{c|}{Std} & \multicolumn{1}{c}{Mean} & \multicolumn{1}{c|}{Std} & \multicolumn{1}{c}{Mean} & \multicolumn{1}{c|}{Std} & \multicolumn{1}{c}{Mean} & \multicolumn{1}{c|}{Std} & \multicolumn{2}{c}{$-$/$\sim$/$+$} \\ \midrule
\bfseries Sobol & 4.48e+01 & 8.24e-01 ($+$) & 3.93e+09 & 1.91e+09 ($+$) & 1.98e+04 & 8.07e+02 ($+$) & 1.49e+01 & 1.02e+00 ($+$) & 2.25e+02 & 5.93e+00 ($+$) & 2.51e+07 & 2.46e+06 ($+$) & \multicolumn{2}{c}{0/4/16} \\
\bfseries NParEGO & 5.06e+01 & 1.65e+00 ($\sim$) & 1.03e+10 & 4.58e+08 ($+$) & 2.14e+04 & 2.55e+02 ($\sim$) & 2.58e+01 & 9.75e-01 ($\sim$) & 2.45e+02 & 1.73e+01 ($+$) & 4.14e+07 & 4.25e+06 ($+$) & \multicolumn{2}{c}{0/13/7} \\
\bfseries TS-TCH & 4.87e+01 & 1.53e+00 ($+$) & 7.83e+09 & 1.55e+09 ($+$) & 2.13e+04 & 3.01e+02 ($\sim$) & 2.15e+01 & 7.63e-01 ($+$) & 2.37e+02 & 9.85e+00 ($+$) & 3.68e+07 & 2.93e+06 ($+$) & \multicolumn{2}{c}{0/5/15} \\
\bfseries JES & 4.57e+01 & 1.97e+00 ($+$) & 1.03e+10 & 6.69e+08 ($+$) & 2.11e+04 & 5.55e+02 ($+$) & 2.36e+01 & 1.48e+00 ($+$) & 2.32e+02 & 1.86e+01 ($+$) & 4.05e+07 & 3.37e+06 ($+$) & \multicolumn{2}{c}{0/9/11} \\
\bfseries NEHVI & 5.09e+01 & 1.99e+00 ($\sim$) & \textbf{1.07e+10} & \textbf{3.32e+08} ($\sim$) & 2.14e+04 & 4.33e+02 ($\sim$) & \textbf{2.68e+01} & \textbf{3.74e-01} ($-$) & \textbf{2.89e+02} & \textbf{1.90e+01} ($-$) & 4.52e+07 & 5.15e+06 ($\sim$) & \multicolumn{2}{c}{3/11/6} \\
\bfseries MORBO & 4.81e+01 & 1.89e+00 ($+$) & 8.23e+09 & 1.49e+09 ($+$) & 2.13e+04 & 4.11e+02 ($\sim$) & 2.08e+01 & 9.21e-01 ($+$) & 2.37e+02 & 1.21e+01 ($+$) & 3.78e+07 & 2.20e+06 ($+$) & \multicolumn{2}{c}{0/5/15} \\
\bfseries qPOTS & 4.85e+01 & 1.57e+00 ($+$) & 8.27e+09 & 1.19e+09 ($+$) & 2.14e+04 & 2.67e+02 ($\sim$) & 2.09e+01 & 8.46e-01 ($+$) & 2.39e+02 & 1.03e+01 ($+$) & 3.87e+07 & 2.54e+06 ($+$) & \multicolumn{2}{c}{0/5/15} \\
\bfseries CTD (ours) & \textbf{5.11e+01} & \textbf{1.08e+00} & 1.06e+10 & 2.83e+08 & \textbf{2.15e+04} & \textbf{1.70e+02} & 2.57e+01 & 6.86e-01 & 2.57e+02 & 1.95e+01 & \textbf{4.52e+07} & \textbf{5.62e+06} & \multicolumn{2}{c}{--/--/--} \\
\bottomrule
\end{tabular}
}
\end{table*}



\paragraph{Noisy Cases.}

Then, we consider noisy settings. Here, we do not include PSL~\citep{lin2022pareto} and MOBO-OSD~\citep{ngo2025moboosd}, as their original studies focus on noiseless settings. 
For EI-based methods, i.e., ParEGO, EHVI, and CTD, we use their noisy variants~\citep{ament2023unexpected,daulton2021parallel}. 
We evaluate the proposed CTD on the same 20 benchmark and real-world problems as in the noiseless setting but with observation noise. 

Table~\ref{tab:noisy_100} shows the HV results (mean and standard deviation) of our method and the seven peer methods on the 20 benchmark and real-world problems. 
Similar to the noiseless setting, CTD demonstrates a clear performance advantage over its competitors. In total, it performs statistically better, equivalently and worse in 85, 52 and 3 out of the 140 pairwise comparisons, respectively. 




\paragraph{High-Dimensional Problems.}
We now investigate how the proposed CTD performs on high-dimensional problems. In this experiment, we include a new MOBO algorithm dedicated to high-dimensional problems,
MORBO~\citep{daulton2022multi}. 
We consider the scalable problem suite DTLZ, which are frequently used in the high-dimensional MOBO literature~\citep{daulton2022multi,rashidi2024cylindrical}.
We set the dimensionality $d=50$ for DTLZ1--DTLZ7 with $m=2$ and $m=3$ objectives. 
Note that since our initial design contains $2(d+1)$ points, i.e., 102 points, 
which already exceeds the budget of 100 evaluations used in the previous experiments, 
we allow a total budget of 200 evaluations, following the practice in~\citep{daulton2020differentiable,daulton2021parallel,konakovic2020diversity}.

The HV results (mean and standard deviation) of each method can be found in Appendix~\ref{appendix:sec:HD}. 
As shown, the advantage of the proposed CTD is even clearer in high-dimensional cases. 
It performs statistically better, equivalently and worse in 111, 8 and 7 out of the 126 pairwise comparisons, respectively. 
This suggests that prioritising convergence before diversity is effective when the search space becomes substantially larger. 
We observe that CTD statistically outperforms the high-dimensional method MORBO on all the problems. One possible reason for the weaker performance of MORBO is that the method may be better suited to settings with generous evaluation budgets, e.g., 2,000 evaluations~\citep{daulton2022multi}.

\begin{figure}[!ht]
    \centering
    \includegraphics[width=\linewidth]{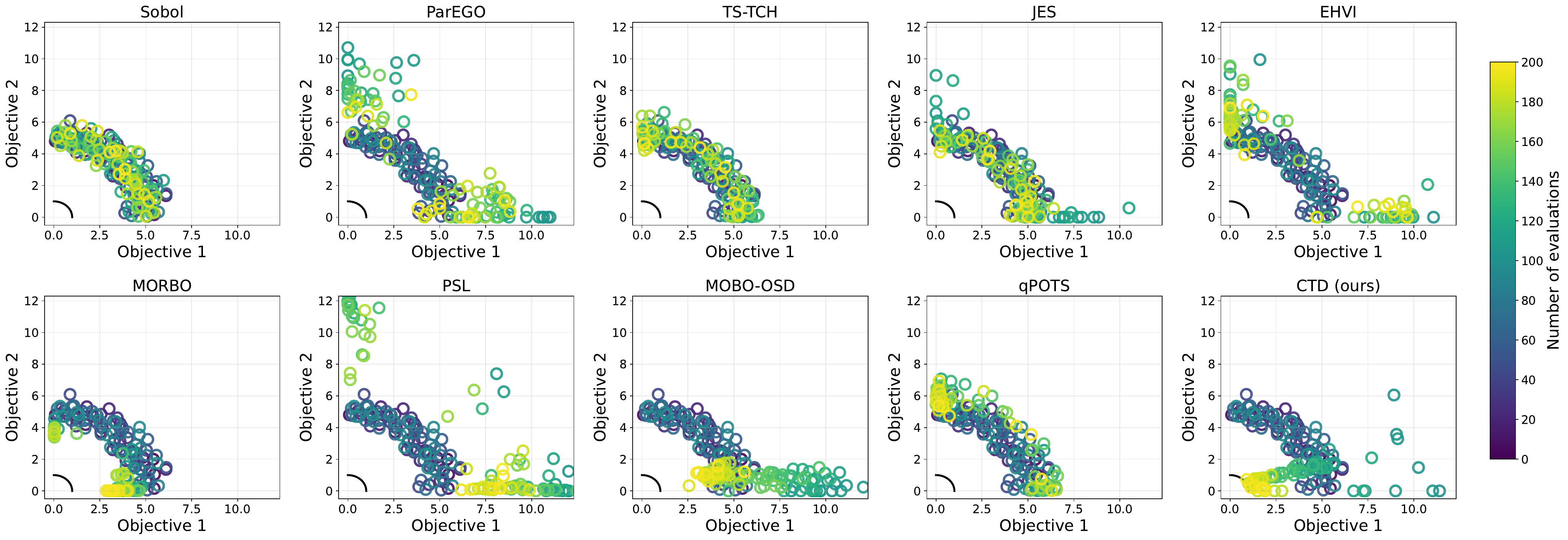}
    \caption{Solutions obtained by the proposed CTD and the nine peer methods on DTLZ2 ($m=2$, $d=50$) under 200 evaluations. All methods start from the same $2(d+1)$ initial points generated by a scrambled Sobol sequence. The black line denotes the Pareto front, and brighter colours indicate solutions obtained at later evaluations.}
    \label{fig:hd}
\end{figure}

To visually understand the results, Figure~\ref{fig:hd} shows the solutions obtained by each method on DTLZ2 ($m=2$, $d=50$), where all methods start from the same $2(d+1)$ initial points generated by a scrambled Sobol sequence. 
As shown in the figure, the proposed CTD is able to approach the Pareto front, while the other methods remain distant from the Pareto front. 
This indicates that the proposed convergence-first strategy is particularly effective in high-dimensional problems under the limited evaluation budget.

\begin{table*}[!ht]
\centering
\caption{The HV results (mean and standard deviation) of our method with six different threshold on the 20 benchmark and real-world problems. 
The method with the best mean HV is highlighted in bold. The symbols ``$-$'', ``$\sim$'' and ``$+$'' indicate that our method CTD is statistically worse than, equivalent to, and better than the competitor, respectively.}
\label{tab:sensitivity}
\resizebox{\textwidth}{!}{
\begin{tabular}{l | ll| ll| ll| ll| ll| ll|ll}
\toprule
\bfseries Method & \multicolumn{2}{c|}{\bfseries DTLZ1 ($m=2$)} & \multicolumn{2}{c|}{\bfseries DTLZ2 ($m=2$)} & \multicolumn{2}{c|}{\bfseries DTLZ3 ($m=2$)} & \multicolumn{2}{c|}{\bfseries DTLZ4 ($m=2$)} & \multicolumn{2}{c|}{\bfseries DTLZ5 ($m=2$)} & \multicolumn{2}{c|}{\bfseries DTLZ6 ($m=2$)} & \multicolumn{2}{c}{\bfseries DTLZ7 ($m=2$)} \\
 & \multicolumn{1}{c}{Mean} & \multicolumn{1}{c|}{Std} & \multicolumn{1}{c}{Mean} & \multicolumn{1}{c|}{Std} & \multicolumn{1}{c}{Mean} & \multicolumn{1}{c|}{Std} & \multicolumn{1}{c}{Mean} & \multicolumn{1}{c|}{Std} & \multicolumn{1}{c}{Mean} & \multicolumn{1}{c|}{Std} & \multicolumn{1}{c}{Mean} & \multicolumn{1}{c|}{Std} & \multicolumn{1}{c}{Mean} & \multicolumn{1}{c}{Std} \\ \midrule
\bfseries CTD ${(10^{-1})}$ & 0.00e+00 & 0.00e+00 ($\sim$) & 1.93e-01 & 1.07e-01 ($+$) & 5.42e+02 & 7.71e+02 ($\sim$) & 2.21e-01 & 8.52e-02 ($+$) & 2.20e-01 & 9.21e-02 ($+$) & \textbf{3.26e-01} & \textbf{4.89e-02} ($\sim$) & 1.73e-01 & 2.61e-02 ($\sim$) \\
\bfseries CTD ${(10^{-2})}$ & 1.92e+00 & 1.05e+01 ($\sim$) & 3.17e-01 & 5.62e-02 ($\sim$) & 4.93e+02 & 1.06e+03 ($\sim$) & 3.16e-01 & 4.19e-02 ($+$) & 3.17e-01 & 4.51e-02 ($+$) & 3.07e-01 & 5.16e-02 ($\sim$) & 1.71e-01 & 2.43e-02 ($\sim$) \\
\bfseries CTD ${(10^{-3})}$ & 1.87e-01 & 8.49e-01 ($\sim$) & 3.44e-01 & 2.32e-02 ($\sim$) & 6.01e+02 & 1.38e+03 ($\sim$) & \textbf{3.55e-01} & \textbf{1.46e-02} ($\sim$) & 3.40e-01 & 2.22e-02 ($\sim$) & 3.00e-01 & 5.33e-02 ($\sim$) & 1.70e-01 & 2.37e-02 ($\sim$) \\
\bfseries CTD ${(10^{-5})}$ & 5.18e-01 & 2.84e+00 ($\sim$) & 3.33e-01 & 3.64e-02 ($\sim$) & \textbf{7.00e+02} & \textbf{9.85e+02} ($\sim$) & 3.46e-01 & 2.06e-02 ($\sim$) & 3.43e-01 & 2.33e-02 ($\sim$) & 3.17e-01 & 6.28e-02 ($\sim$) & \textbf{1.75e-01} & \textbf{2.73e-02} ($\sim$) \\
\bfseries CTD ${(10^{-6})}$ & 0.00e+00 & 0.00e+00 ($\sim$) & 2.90e-01 & 4.78e-02 ($+$) & 6.96e+02 & 8.26e+02 ($\sim$) & 3.02e-01 & 3.33e-02 ($+$) & 2.90e-01 & 5.75e-02 ($+$) & 3.15e-01 & 5.20e-02 ($\sim$) & 1.62e-01 & 1.20e-02 ($\sim$) \\
\bfseries CTD ${(10^{-4})}$ & \textbf{6.68e+00} & \textbf{2.43e+01} & \textbf{3.46e-01} & \textbf{2.09e-02} & 4.17e+02 & 8.64e+02 & 3.52e-01 & 1.46e-02 & \textbf{3.49e-01} & \textbf{1.95e-02} & 3.19e-01 & 4.51e-02 & 1.71e-01 & 2.43e-02 \\
\bottomrule
\toprule
\bfseries Method & \multicolumn{2}{c|}{\bfseries DTLZ1 ($m=3$)} & \multicolumn{2}{c|}{\bfseries DTLZ2 ($m=3$)} & \multicolumn{2}{c|}{\bfseries DTLZ3 ($m=3$)} & \multicolumn{2}{c|}{\bfseries DTLZ4 ($m=3$)} & \multicolumn{2}{c|}{\bfseries DTLZ5 ($m=3$)} & \multicolumn{2}{c|}{\bfseries DTLZ6 ($m=3$)} & \multicolumn{2}{c}{\bfseries DTLZ7 ($m=3$)} \\
 & \multicolumn{1}{c}{Mean} & \multicolumn{1}{c|}{Std} & \multicolumn{1}{c}{Mean} & \multicolumn{1}{c|}{Std} & \multicolumn{1}{c}{Mean} & \multicolumn{1}{c|}{Std} & \multicolumn{1}{c}{Mean} & \multicolumn{1}{c|}{Std} & \multicolumn{1}{c}{Mean} & \multicolumn{1}{c|}{Std} & \multicolumn{1}{c}{Mean} & \multicolumn{1}{c|}{Std} & \multicolumn{1}{c}{Mean} & \multicolumn{1}{c}{Std} \\ \midrule
\bfseries CTD ${(10^{-1})}$ & \textbf{9.86e+03} & \textbf{5.85e+03} ($-$) & 5.05e-02 & 7.32e-02 ($+$) & \textbf{2.12e+06} & \textbf{7.00e+05} ($-$) & 5.98e-02 & 5.00e-02 ($+$) & 2.30e-02 & 2.47e-02 ($+$) & 9.28e-02 & 1.86e-02 ($\sim$) & \textbf{2.26e-01} & \textbf{1.14e-02} ($\sim$) \\
\bfseries CTD ${(10^{-2})}$ & 6.58e+03 & 6.06e+03 ($\sim$) & 2.21e-01 & 7.10e-02 ($+$) & 1.63e+06 & 7.83e+05 ($\sim$) & 2.37e-01 & 9.99e-02 ($+$) & 8.08e-02 & 1.95e-02 ($+$) & 9.51e-02 & 1.69e-02 ($\sim$) & 2.25e-01 & 1.16e-02 ($\sim$) \\
\bfseries CTD ${(10^{-3})}$ & 6.24e+03 & 7.01e+03 ($\sim$) & 3.31e-01 & 9.33e-02 ($\sim$) & 1.59e+06 & 9.17e+05 ($\sim$) & 3.03e-01 & 7.64e-02 ($+$) & 9.42e-02 & 1.64e-02 ($\sim$) & 9.94e-02 & 1.76e-02 ($\sim$) & 2.26e-01 & 1.14e-02 ($\sim$) \\
\bfseries CTD ${(10^{-5})}$ & 1.67e+03 & 2.81e+03 ($\sim$) & 3.73e-01 & 7.25e-02 ($\sim$) & 1.39e+06 & 8.29e+05 ($\sim$) & 3.50e-01 & 6.30e-02 ($\sim$) & 9.59e-02 & 2.03e-02 ($\sim$) & 8.90e-02 & 2.17e-02 ($\sim$) & 2.24e-01 & 1.58e-02 ($\sim$) \\
\bfseries CTD ${(10^{-6})}$ & 3.93e+03 & 6.34e+03 ($\sim$) & 3.67e-01 & 6.17e-02 ($\sim$) & 1.42e+06 & 7.98e+05 ($\sim$) & \textbf{3.68e-01} & \textbf{7.43e-02} ($\sim$) & 8.71e-02 & 2.27e-02 ($+$) & 8.71e-02 & 2.52e-02 ($+$) & 1.97e-01 & 3.17e-02 ($+$) \\
\bfseries CTD ${(10^{-4})}$ & 4.29e+03 & 5.30e+03 & \textbf{3.73e-01} & \textbf{7.95e-02} & 1.58e+06 & 8.91e+05 & 3.60e-01 & 9.01e-02 & \textbf{1.02e-01} & \textbf{1.30e-02} & \textbf{1.01e-01} & \textbf{1.73e-02} & 2.22e-01 & 1.90e-02 \\
\bottomrule
\toprule
\bfseries Method & \multicolumn{2}{c|}{\bfseries Four bar truss design} & \multicolumn{2}{c|}{\bfseries Pressure vessel design} & \multicolumn{2}{c|}{\bfseries Hatch cover design} & \multicolumn{2}{c|}{\bfseries Vehicle safety} & \multicolumn{2}{c|}{\bfseries Car side impact} & \multicolumn{2}{c|}{\bfseries LPA} & \multicolumn{2}{c}{\bfseries Summary} \\
 & \multicolumn{1}{c}{Mean} & \multicolumn{1}{c|}{Std} & \multicolumn{1}{c}{Mean} & \multicolumn{1}{c|}{Std} & \multicolumn{1}{c}{Mean} & \multicolumn{1}{c|}{Std} & \multicolumn{1}{c}{Mean} & \multicolumn{1}{c|}{Std} & \multicolumn{1}{c}{Mean} & \multicolumn{1}{c|}{Std} & \multicolumn{1}{c}{Mean} & \multicolumn{1}{c|}{Std} & \multicolumn{2}{c}{$-$/$\sim$/$+$} \\ \midrule
\bfseries CTD ${(10^{-1})}$ & 5.41e+01 & 3.30e-02 ($\sim$) & 1.08e+10 & 1.90e+08 ($\sim$) & 2.18e+04 & 2.55e+01 ($\sim$) & 2.59e+01 & 7.70e-01 ($\sim$) & 3.09e+02 & 1.16e+01 ($\sim$) & 4.54e+07 & 4.07e+06 ($\sim$) & \multicolumn{2}{c}{2/12/6} \\
\bfseries CTD ${(10^{-2})}$ & 5.41e+01 & 4.12e-02 ($\sim$) & 1.08e+10 & 1.86e+08 ($\sim$) & \textbf{2.18e+04} & \textbf{2.05e+01} ($\sim$) & 2.58e+01 & 9.54e-01 ($\sim$) & \textbf{3.12e+02} & \textbf{1.25e+01} ($\sim$) & 4.61e+07 & 3.80e+06 ($\sim$) & \multicolumn{2}{c}{0/15/5} \\
\bfseries CTD ${(10^{-3})}$ & \textbf{5.41e+01} & \textbf{3.02e-02} ($\sim$) & 1.09e+10 & 1.24e+08 ($\sim$) & 2.18e+04 & 2.18e+01 ($\sim$) & 2.55e+01 & 1.64e+00 ($\sim$) & 2.80e+02 & 9.56e+01 ($\sim$) & 4.51e+07 & 4.29e+06 ($\sim$) & \multicolumn{2}{c}{0/19/1} \\
\bfseries CTD ${(10^{-5})}$ & 5.40e+01 & 6.27e-02 ($+$) & 1.09e+10 & 1.77e+08 ($\sim$) & 2.18e+04 & 1.88e+01 ($\sim$) & \textbf{2.60e+01} & \textbf{7.75e-01} ($\sim$) & 3.10e+02 & 1.49e+01 ($\sim$) & \textbf{4.64e+07} & \textbf{3.99e+06} ($\sim$) & \multicolumn{2}{c}{0/19/1} \\
\bfseries CTD ${(10^{-6})}$ & 5.30e+01 & 2.77e+00 ($+$) & \textbf{1.09e+10} & \textbf{1.40e+08} ($\sim$) & 2.18e+04 & 4.72e+01 ($\sim$) & 2.46e+01 & 2.18e+00 ($+$) & 2.98e+02 & 2.02e+01 ($\sim$) & 4.46e+07 & 4.44e+06 ($\sim$) & \multicolumn{2}{c}{0/12/8} \\
\bfseries CTD ${(10^{-4})}$ & 5.41e+01 & 2.82e-02 & 1.09e+10 & 1.79e+08 & 2.18e+04 & 2.05e+01 & 2.60e+01 & 6.38e-01 & 3.05e+02 & 1.55e+01 & 4.62e+07 & 3.71e+06 & \multicolumn{2}{c}{--/--/--} \\
\bottomrule
\end{tabular}
}
\end{table*}

It is worth mentioning that the results obtained by the proposed method can echo recent findings in single-objective Bayesian optimisation, where standard Gaussian processes with scaled priors can perform well in high-dimensional spaces~\citep{hvarfner2024vanilla,papenmeier2025understanding,xu2025standard} even without using high-dimensional handling techniques~\citep{binois2022survey,gonzalez2024survey,papenmeier2023bounce,santoni2024comparison,wang2016bayesian,hoang2025high,doumont2025we} (e.g., variable selection~\citep{eriksson2021high}, additive models~\citep{delbridge2020randomly,han2021high,wang2018batched,ziomek2023random}, embeddings~\citep{antonov2022high,letham2020re,raponi2020high}, and trust regions~\citep{daulton2022multi,diouane2023trego,eriksson2019scalable}).


\subsection{Sensitivity Analysis}\label{sec:sensitivity}

Within the proposed method CTD, the threshold $\kappa$ is used to control when the search switches to the diversity-focused stage by comparing it with the maximum EI value.
A large $\kappa$ may result in solutions that are still far from the Pareto front before the switch, whereas a small $\kappa$ may lead to unnecessary exploration after the search has already converged. 

In this work, we set $\kappa=10^{-4}$, as suggested by~\citep{nguyen2017regret}. The results show that this setting allows the search to obtain well-converged solutions and can switch to diversity-focused search without unnecessary exploration (see Figure~\ref{fig:dtlz2-pf-comparison}). We further now evaluate the sensitivity of $\kappa$ and set $\kappa\in\{10^{-1},10^{-2},10^{-3},10^{-5},10^{-6}\}$. Table~\ref{tab:sensitivity} shows the HV results (mean and standard deviation) of our method with six  different thresholds on the 20 benchmark and real-world problems.

As can be seen, CTD is relatively insensitive to $\kappa$ within the range from $10^{-3}$ to $10^{-5}$. Compared with the default setting $\kappa=10^{-4}$, both $\kappa=10^{-3}$ and $\kappa=10^{-5}$ achieve statistically equivalent results on 19 out of the 20 problems. By contrast, the more extreme settings generally lead to poorer performance. Setting $\kappa$ to $10^{-1}$ or $10^{-2}$ may trigger the diversity-focused stage while the solutions are still far from the Pareto front, resulting in statistically worse performance on six and five problems, respectively. Meanwhile, setting $\kappa=10^{-6}$ may consume unnecessary evaluations after the search has already converged, resulting in statistically worse performance on eight problems.

\vspace{-5pt}
\subsection{Using an Alternative Acquisition Function}\label{sec:dif_acf}

\begin{table*}[t]
\centering
\caption{The HV results (mean and standard deviation) of the two methods on the 20 benchmark and real-world problems on 30 independent runs. 
The method with the best mean HV is highlighted in bold. The symbols ``$-$'', ``$\sim$'' and ``$+$'' indicate that CTD$_{tch}$ is statistically better than, equivalent to, and worse than EHVI-based CTD, respectively.}
\label{tab:noiseless_100_acf}
\resizebox{\textwidth}{!}{
\begin{tabular}{l | ll| ll| ll| ll| ll| ll|ll}
\toprule
\bfseries Method & \multicolumn{2}{c|}{\bfseries DTLZ1 ($m=2$)} & \multicolumn{2}{c|}{\bfseries DTLZ2 ($m=2$)} & \multicolumn{2}{c|}{\bfseries DTLZ3 ($m=2$)} & \multicolumn{2}{c|}{\bfseries DTLZ4 ($m=2$)} & \multicolumn{2}{c|}{\bfseries DTLZ5 ($m=2$)} & \multicolumn{2}{c|}{\bfseries DTLZ6 ($m=2$)} & \multicolumn{2}{c}{\bfseries DTLZ7 ($m=2$)} \\
 & \multicolumn{1}{c}{Mean} & \multicolumn{1}{c|}{Std} & \multicolumn{1}{c}{Mean} & \multicolumn{1}{c|}{Std} & \multicolumn{1}{c}{Mean} & \multicolumn{1}{c|}{Std} & \multicolumn{1}{c}{Mean} & \multicolumn{1}{c|}{Std} & \multicolumn{1}{c}{Mean} & \multicolumn{1}{c|}{Std} & \multicolumn{1}{c}{Mean} & \multicolumn{1}{c|}{Std} & \multicolumn{1}{c}{Mean} & \multicolumn{1}{c}{Std} \\ \midrule
\bfseries CTD$_{tch}$ & 0.00e+00 & 0.00e+00 ($\sim$) & \textbf{3.55e-01} & \textbf{1.22e-02} ($\sim$) & \textbf{5.07e+02} & \textbf{6.87e+02} ($\sim$) & \textbf{3.56e-01} & \textbf{1.20e-02} ($\sim$) & \textbf{3.51e-01} & \textbf{1.52e-02} ($\sim$) & \textbf{3.53e-01} & \textbf{6.73e-02} ($-$) & \textbf{3.75e-01} & \textbf{2.16e-01} ($-$) \\
\bfseries CTD & \textbf{6.68e+00} & \textbf{2.43e+01} & 3.46e-01 & 2.09e-02 & 4.17e+02 & 8.64e+02 & 3.52e-01 & 1.46e-02 & 3.49e-01 & 1.95e-02 & 3.19e-01 & 4.51e-02 & 1.71e-01 & 2.43e-02 \\
\bottomrule
\toprule
\bfseries Method & \multicolumn{2}{c|}{\bfseries DTLZ1 ($m=3$)} & \multicolumn{2}{c|}{\bfseries DTLZ2 ($m=3$)} & \multicolumn{2}{c|}{\bfseries DTLZ3 ($m=3$)} & \multicolumn{2}{c|}{\bfseries DTLZ4 ($m=3$)} & \multicolumn{2}{c|}{\bfseries DTLZ5 ($m=3$)} & \multicolumn{2}{c|}{\bfseries DTLZ6 ($m=3$)} & \multicolumn{2}{c}{\bfseries DTLZ7 ($m=3$)} \\
 & \multicolumn{1}{c}{Mean} & \multicolumn{1}{c|}{Std} & \multicolumn{1}{c}{Mean} & \multicolumn{1}{c|}{Std} & \multicolumn{1}{c}{Mean} & \multicolumn{1}{c|}{Std} & \multicolumn{1}{c}{Mean} & \multicolumn{1}{c|}{Std} & \multicolumn{1}{c}{Mean} & \multicolumn{1}{c|}{Std} & \multicolumn{1}{c}{Mean} & \multicolumn{1}{c|}{Std} & \multicolumn{1}{c}{Mean} & \multicolumn{1}{c}{Std} \\ \midrule
\bfseries CTD$_{tch}$ & \textbf{4.30e+03} & \textbf{7.83e+03} ($\sim$) & \textbf{4.68e-01} & \textbf{5.62e-02} ($-$) & \textbf{1.60e+06} & \textbf{8.57e+05} ($\sim$) & \textbf{4.78e-01} & \textbf{5.39e-02} ($-$) & \textbf{1.05e-01} & \textbf{1.43e-02} ($\sim$) & 9.86e-02 & 2.38e-02 ($\sim$) & \textbf{4.58e-01} & \textbf{2.68e-01} ($\sim$) \\
\bfseries CTD & 4.29e+03 & 5.30e+03 & 3.73e-01 & 7.95e-02 & 1.58e+06 & 8.91e+05 & 3.60e-01 & 9.01e-02 & 1.02e-01 & 1.30e-02 & \textbf{1.01e-01} & \textbf{1.73e-02} & 2.22e-01 & 1.90e-02 \\
\bottomrule
\toprule
\bfseries Method & \multicolumn{2}{c|}{\bfseries Four bar truss design} & \multicolumn{2}{c|}{\bfseries Pressure vessel design} & \multicolumn{2}{c|}{\bfseries Hatch cover design} & \multicolumn{2}{c|}{\bfseries Vehicle safety} & \multicolumn{2}{c|}{\bfseries Car side impact} & \multicolumn{2}{c|}{\bfseries LPA} & \multicolumn{2}{c}{\bfseries Summary} \\
 & \multicolumn{1}{c}{Mean} & \multicolumn{1}{c|}{Std} & \multicolumn{1}{c}{Mean} & \multicolumn{1}{c|}{Std} & \multicolumn{1}{c}{Mean} & \multicolumn{1}{c|}{Std} & \multicolumn{1}{c}{Mean} & \multicolumn{1}{c|}{Std} & \multicolumn{1}{c}{Mean} & \multicolumn{1}{c|}{Std} & \multicolumn{1}{c}{Mean} & \multicolumn{1}{c|}{Std} & \multicolumn{2}{c}{$-$/$\sim$/$+$} \\ \midrule
\bfseries CTD$_{tch}$ & 5.13e+01 & 1.51e+00 ($+$) & \textbf{1.10e+10} & \textbf{3.73e+07} ($-$) & 2.18e+04 & 3.11e+01 ($+$) & 2.59e+01 & 1.61e+00 ($\sim$) & \textbf{3.26e+02} & \textbf{1.09e+01} ($-$) & 4.13e+07 & 3.45e+06 ($+$) & \multicolumn{2}{c}{6/11/3} \\
\bfseries CTD & \textbf{5.41e+01} & \textbf{2.82e-02} & 1.09e+10 & 1.79e+08 & \textbf{2.18e+04} & \textbf{2.05e+01} & \textbf{2.60e+01} & \textbf{6.38e-01} & 3.05e+02 & 1.55e+01 & \textbf{4.62e+07} & \textbf{3.71e+06} & \multicolumn{2}{c}{--/--/--} \\
\bottomrule
\end{tabular}
}
\end{table*}

In the proposed CTD framework, we employ EHVI as the acquisition function during the diversity-focused search. 
In this section, we want to investigate whether other acquisition functions also work.
We now consider EI with random augmented Tchebycheff scalarisations (used in ParEGO~\cite{knowles2006parego}) in the diversity-focused search, denoted as CTD$_{tch}$. 

Table~\ref{tab:noiseless_100_acf} shows the results (mean and standard deviation) of the HV obtained by the proposed CTD with different acquisition functions (EI based on HV and random scalarisations) in the diversity-focused search across 20 benchmark and real-world problems under the noiseless setting. 
Interestingly, CTD with random scalarisations (CTD$_{tch}$) is highly competitive, performing statistically better, equivalently, and worse than the original CTD on 6, 11, and 3 out of the 20 problems, respectively.
Notably, CTD$_{tch}$ improves the performance on some problems (e.g., DTLZ7) where the EHVI-based CTD performs relatively poorly (see Table~\ref{tab:noiseless_100}). This suggests that CTD itself does not restrict global exploration, while a more suitable acquisition function can significantly improve its performance. 
Due to space limitation, the HV results of CTD$_{tch}$ under noisy and high-dimensional settings are placed in Appendix~\ref{appendix:sec:acf}. 
They are consistent with those in the noiseless setting: CTD$_{tch}$ is statistically better than, equivalent to, and worse than the original CTD on 4, 11, and 5 out of the 20 noisy problems, respectively, and on 7, 4, and 3 out of the 14 high-dimensional problems, respectively.



\vspace{-5pt}
\section{Conclusion}\label{sec:con}

This work presented a new MOBO framework, CTD, that prioritises convergence before diversity, rather than balancing both throughout the search. We evaluated CTD on a range of benchmark and real-world problems, covering noiseless, noisy, and high-dimensional settings. The results show that CTD performs effectively across these scenarios. We used EHVI as the default acquisition function in the diversity-focused stage, and found that using EI based on random augmented Tchebycheff scalarisations (i.e., ParEGO) can achieve even better performance. 
Other acquisition functions for the diversity-focused stage could be further explored in future work. 
For the convergence-focused stage and the switching criterion, we used conventional methods: EI with the augmented Tchebycheff scalarisation and an EI-based stopping rule, respectively.
Future work could explore other methods, including alternative acquisition functions (e.g., UCB~\citep{lai1985asymptotically}) and scalarisations (e.g., smooth Tchebycheff scalarisation~\citep{lin2024smooth}) for the convergence-focused search, as well as other BO stopping rules~\citep{lorenz2015stopping,frazier2008knowledge,xie2025cost,makarova2022automatic,ishibashi2023stopping,wilson2024stopping} for determining the switching time.

\bibliographystyle{unsrt}

\bibliography{refs}


\clearpage

\appendix

\begin{center}
\hrule height 4pt
\vskip 0.25in
\vskip -\parskip
    {\LARGE\bf  Appendix to:\\[2ex] \papertitle}
\vskip 0.29in
\vskip -\parskip
\hrule height 1pt
\vskip 0.2in%
\end{center}

\section{Multi-Objective Bayesian Optimisation (MOBO)}\label{appdx:sec:BO}
MOBO consists of two main steps, i.e., training $m$ Gaussian process models based on the observed solutions and optimising an acquisition function $\alpha: \mathcal{X} \to \mathbb{R}$ to select a solution for evaluation. 
In this work, we model each objective with an independent Gaussian process $f_i \sim \mathcal{GP}(m_i(\bm x),k_i(\bm x,\bm x'))$, where $m_i: \mathcal{X} \to \mathbb{R}$ is the $i$th mean function, and $k_i(\cdot,\cdot): \mathcal{X} \times \mathcal{X} \to \mathbb{R}$ is the $i$th covariance function. 
We use the notation $K(\mathrm{A}, \mathrm{B})$ to represent the covariance matrix at all pairs of solutions in set $\mathrm{A}$ and in set $\mathrm{B}$. 
Given $n$ observed points $\mathcal{D}^{(n)}= \{(\bm x^{(t)},\bm y^{(t)})\}_{t=1}^{n}$, where $\bm y^{(t)} = \bm{f}(\bm x^{(t)}) + \bm\zeta^{(t)}$ and the noise $\bm \zeta^{(t)} \sim \mathcal{N}(\bm 0, \text{diag} (\bm \sigma_\zeta^2))$, the posterior distribution of the $i$th objective at a new location $\bm x$ is Gaussian: 
\begin{equation}
	f_i(\bm x)\mid\mathcal{D}^{(n)} \sim \mathcal{N}(\mu_i(\bm x),\sigma_i^2(\bm x))   
\end{equation}
\begin{equation}
	\mu_i(\bm x)=K(\bm x,X^{(n)})(K(X^{(n)},X^{(n)})+\sigma_{\zeta_i}^2\mathbf{I})^{-1}Y_i^{(n)}
\end{equation}
\begin{equation}
	\sigma_i^2(\bm x)=K(\bm x,\bm x)-K(\bm x,X^{(n)})(K(X^{(n)},X^{(n)})+\sigma_{\zeta_i}^2\mathbf{I})^{-1}K(X^{(n)},\bm x) 
\end{equation}
where $\mu_i(\bm x)$ and $\sigma_i^2(\bm x)$ are the mean and variance at $\bm x$, respectively; 
$X^{(n)}=(\bm x^{(1)},\dots,\bm x^{(n)})\in \mathbb{R}^{n\times d}$ and $Y_i^{(n)}=(y_i^{(1)},\dots,y_i^{(n)})\in \mathbb{R}^{n}$ are the matrix of evaluated solutions and the corresponding vector of $y$ values, respectively; 
$\sigma_{\zeta_i}^2$ is the variance of the observation noise $\zeta_i \sim \mathcal{N}(0, \sigma_{\zeta_i}^2)$, and corresponds to the $i$-th diagonal entry of the noise covariance matrix $\text{diag}(\bm{\sigma}_\zeta^2)$; 

\section{Experiment Settings}\label{appdx:sec:setting}

\subsection{Implementation details}\label{appdx:sec:implementation}

The computational studies were performed on a Red Hat Enterprise Linux 8.8 system, operating on a 64-bit x86 CPU architecture. The computing cluster utilised Intel Xeon Platinum 8360Y processors running at 2.40 GHz. 
The implementation details are as follows. 


\paragraph{CTD.} We implement CTD based on the open-source Python framework BoTorch~\cite{balandat2020botorch}, which builds on GPyTorch~\cite{gardner2018gpytorch} for Gaussian process modelling and PyTorch~\cite{paszke2019pytorch} for automatic differentiation.
For the convergence-focused search, we use LogEI~\citep{ament2023unexpected} based on the augmented Tchebycheff scalarisation~\citep{miettinen1999nonlinear} with a fixed weight vector $(\frac{1}{m},\dots,\frac{1}{m}) \in \mathbb{R}^m$. 
For the diversity-focused search, we use LogEHVI~\citep{ament2023unexpected} as the main implementation and LogEI~\citep{ament2023unexpected} with random scalarisations as an alternative diversity-focused variant. 
In noisy cases, we replace LogEI with LogNEI and LogEHVI with LogNEHVI~\citep{ament2023unexpected,daulton2021parallel}. 
For the switching criterion, we set $\kappa=10^{-4}$ (i.e., the maximum EI value is smaller than $10^{-4}$), 
according to~\citep{nguyen2017regret}. Since LogEI~\citep{ament2023unexpected} is used in the implementation, the corresponding threshold is applied on the log scale, i.e., $\log(\kappa)$.

\paragraph{ParEGO~\citep{knowles2006parego} and NParEGO~\citep{daulton2021parallel}.} 

We implement ParEGO and NParEGO based on the implementation provided by BoTorch.\footnote{\href{https://botorch.readthedocs.io/en/latest/_modules/botorch/acquisition/monte_carlo.html\#qExpectedImprovement}{EI}, 
\href{https://botorch.readthedocs.io/en/latest/_modules/botorch/acquisition/monte_carlo.html\#qNoisyExpectedImprovement}{NEI}, 
and the \href{https://botorch.org/docs/tutorials/multi_objective_bo/}{BoTorch multi-objective tutorial}.} 
Both methods use the augmented Tchebycheff scalarisation function and their log version, following the recommendation in~\citep{ament2023unexpected}. 

\paragraph{TS-TCH~\citep{paria2020flexible}.} We implement TS-TCH based on the implementation provided by BoTorch.\footnote{\href{https://botorch.readthedocs.io/en/latest/_modules/botorch/generation/sampling.html\#MaxPosteriorSampling}{TS} and 
\href{https://botorch.org/docs/tutorials/multi_objective_bo/}{BoTorch multi-objective tutorial}} 
The augmented Tchebycheff scalarisation function is used. After converting to single-objective optimisation problem, Thompson sampling is used as the acquisition function. We draw a sample from the joint posterior over a discrete set of $1000d$ points sampled from a scrambled Sobol sequence, suggested by~\citep{daulton2020differentiable}.

\paragraph{JES~\citep{tu2022joint}.} We implement JES using the open-source implementation provided by the authors\footnote{\href{https://github.com/benmltu/JES}{JES implementation}}.

\paragraph{EHVI~\citep{daulton2020differentiable} and NEHVI~\citep{daulton2021parallel}.} We implement EHVI and NEHVI based on the implementation provided by BoTorch.\footnote{\href{https://botorch.readthedocs.io/en/latest/acquisition.html\#botorch.acquisition.multi_objective.logei.qLogExpectedHypervolumeImprovement}{EHVI},
\href{https://botorch.readthedocs.io/en/latest/acquisition.html\#botorch.acquisition.multi_objective.logei.qLogNoisyExpectedHypervolumeImprovement}{NEHVI}, and
\href{https://botorch.org/docs/tutorials/multi_objective_bo/}{BoTorch multi-objective tutorial}}. 
For both methods, we use their log versions, following the recommendation of~\citep{ament2023unexpected}.

\paragraph{MORBO~\citep{daulton2022multi}.} We implement MORBO using the open-source implementation provided by the authors\footnote{\href{https://github.com/facebookresearch/morbo}{MORBO implementation}} and set the batch size to one.

\paragraph{PSL~\citep{lin2022pareto}.} We implement PSL using the open-source implementation provided by the authors\footnote{\href{https://github.com/Xi-L/PSL-MOBO}{PSL implementation}} and set the batch size to one.

\paragraph{MOBO-OSD~\citep{ngo2025moboosd}.} We implement MOBO-OSD using the open-source implementation provided by the authors\footnote{\href{https://github.com/LamNgo1/mobo-osd}{MOBO-OSD implementation}} and set the batch size to one.

\paragraph{qPOTS~\citep{renganathan2025qpots}.} We implement qPOTS using the open-source implementation provided by the authors\footnote{\href{https://github.com/csdlpsu/qpots}{qPOTS implementation}} and set the batch size to one. 

For all the problems, we normalise the input variables and standardise the objective values before training Gaussian processes. 
We assume an independent surrogate model for each objective, using a constant mean function and a Matérn 5/2 ARD kernel with a dimension-scaled prior (see \url{https://github.com/meta-pytorch/botorch/releases/tag/v0.12.0}).

\subsection{Problem Details}\label{appendix:subsec:problems}

The details of the benchmark problems and real-world problems are given in the following. 


\paragraph{DTLZ1--DTLZ7.} DTLZ1--DTLZ7 are scalable multi-objective benchmark problems. Their mathematical formulations are provided in~\cite{deb2005scalable}. 

\paragraph{Four Bar Truss Design.}

The four-bar truss design problem aims to optimise a lightweight truss structure while controlling its structural displacement~\citep{cheng1999generalized,tanabe2020easy}. 
It involves $m=2$ objectives with $d=4$ continuous design variables. 
The first objective is related to the structural volume, while the second objective measures the displacement of the truss under load. 
This problem reflects the trade-off between reducing material usage and maintaining structural stiffness.
The problem is formulated as:
\[
\begin{aligned}
\min_{\bm{x}} \quad 
f_1(\bm{x}) &= L\left(2x_1+\sqrt{2}x_2+\sqrt{x_3}+x_4\right), \\
f_2(\bm{x}) &= \frac{FL}{E}
\left(
\frac{2}{x_1}
+\frac{2\sqrt{2}}{x_2}
-\frac{2\sqrt{2}}{x_3}
+\frac{2}{x_4}
\right),
\end{aligned}
\]
where $F=10$, $E=2\times 10^5$, and $L=200$. 
The decision variables are bounded by
\[
x_1, x_4 \in [1,3], \qquad 
x_2, x_3 \in [\sqrt{2},3].
\]

\paragraph{Pressure Vessel Design.}

The pressure vessel design problem aims to optimise the design of a cylindrical pressure vessel subject to engineering requirements on thickness, radius, length, and stress-related constraints~\citep{kannan1994augmented,tanabe2020easy}. 
It involves $m=2$ objectives with $d=4$ design variables. 
The objectives are associated with minimising the manufacturing cost and the overall constraint violation. 
This problem reflects the trade-off between economical design and feasibility under pressure-vessel safety requirements.
The objective functions are:
\[
\begin{aligned}
f_1(\bm{x}) ={}&
0.6224\tilde{x}_1x_3x_4
+1.7781\tilde{x}_2x_3^2
+3.1661\tilde{x}_1^2x_4
+19.84\tilde{x}_1^2x_3,\\
f_2(\bm{x}) ={}&
\max(0,-g_1(\bm{x}))
+\max(0,-g_2(\bm{x}))
+\max(0,-g_3(\bm{x})),
\end{aligned}
\]
where
\[
\tilde{x}_1 = 0.0625\,\mathrm{round}(x_1), 
\qquad 
\tilde{x}_2 = 0.0625\,\mathrm{round}(x_2),
\]
and
\[
\begin{aligned}
g_1(\bm{x}) &= \tilde{x}_1 - 0.0193x_3,\\
g_2(\bm{x}) &= \tilde{x}_2 - 0.00954x_3,\\
g_3(\bm{x}) &= \pi x_3^2x_4 + \frac{4}{3}\pi x_3^3 - 1296000.
\end{aligned}
\]
The decision variables are bounded by
\[
x_1,x_2 \in [1,100], \qquad 
x_3 \in [10,200], \qquad 
x_4 \in [10,240].
\]

\paragraph{Hatch Cover Design.}

The hatch cover design problem considers the structural design of a hatch cover under loading conditions~\citep{amir1989nonlinear,tanabe2020easy}. 
It involves $m=2$ objectives with $d=2$ continuous design variables. 
The objectives are related to minimising the structural weight and controlling the deflection of the hatch cover. 
This problem captures the trade-off between lightweight structural design and mechanical reliability.
The objective functions are:
\[
\begin{aligned}
f_1(\bm{x}) &= x_1 + 120x_2,\\
f_2(\bm{x}) &= \sum_{i=1}^{4}\max(0,-g_i(\bm{x})),
\end{aligned}
\]
where
\[
\begin{aligned}
g_1(\bm{x}) &= 1-\frac{\sigma_b}{\sigma_{b,\max}},\\
g_2(\bm{x}) &= 1-\frac{\tau}{\tau_{\max}},\\
g_3(\bm{x}) &= 1-\frac{\delta}{\delta_{\max}},\\
g_4(\bm{x}) &= 1-\frac{\sigma_b}{\sigma_k},
\end{aligned}
\]
with
\[
\begin{aligned}
\sigma_k &= \frac{Ex_1^2}{100},\\
\sigma_b &= \frac{4500}{x_1x_2},\\
\tau &= \frac{1800}{x_2},\\
\delta &= \frac{56.2\times 10000}{Ex_1x_2^2}.
\end{aligned}
\]
Here, $E=700000$, $\sigma_{b,\max}=700$, $\tau_{\max}=450$, and $\delta_{\max}=1.5$. 
The decision variables are bounded by
\[
x_1 \in [0.5,4], 
\qquad 
x_2 \in [0.5,50].
\]

\paragraph{Vehicle Safety Design.}

The vehicle safety design problem aims to improve crashworthiness performance while reducing the vehicle weight~\citep{liao2008multiobjective,tanabe2020easy}. 
It involves $m=3$ objectives with $d=5$ continuous design variables. 
The objectives include vehicle weight and safety-related crashworthiness responses, such as acceleration and structural intrusion measures. 
This problem reflects the trade-off between lightweight design and passenger safety in vehicle crashworthiness optimisation.
The objective functions are:
\[
\begin{aligned}
f_1(\bm{x}) ={}&
1640.2823
+2.3573285x_1
+2.3220035x_2
+4.5688768x_3
+7.7213633x_4
+4.4559504x_5,\\
f_2(\bm{x}) ={}&
6.5856
+1.15x_1
-1.0427x_2
+0.9738x_3
+0.8364x_4
-0.3695x_1x_4 \\
&+0.0861x_1x_5
+0.3628x_2x_4
-0.1106x_1^2
-0.3437x_3^2
+0.1764x_4^2,\\
f_3(\bm{x}) ={}&
-0.0551
+0.0181x_1
+0.1024x_2
+0.0421x_3
-0.0073x_1x_2
+0.024x_2x_3 \\
&-0.0118x_2x_4
-0.0204x_3x_4
-0.008x_3x_5
-0.0241x_2^2
+0.0109x_4^2.
\end{aligned}
\]
The decision variables are bounded by
\[
x_i \in [1,3], \qquad i=1,\ldots,5.
\]

\paragraph{Car Side Impact Design.} 

The car side-impact problem aims to minimise vehicle weight while satisfying safety constraints related to occupant injury and structural response~\cite{jain2013evolutionary}. 
It involves $m=4$ objectives with $d=7$ variables. 
The problem reflects the trade-off between safety performance and weight under uncertainty. 
The mathematical formulations are given as follow:

\begin{align*}
f_1(\bm{x}) &= 1.98 + 4.9x_1 + 6.67x_2 + 6.98x_3 + 4.01x_4 + 1.78x_5 + 10^{-5}x_6 + 2.73x_7 \\
f_2(\bm{x}) &= 4.72 - 0.5x_4 - 0.19x_2x_3 \\
f_3(\bm{x}) &= 0.5 \left( V_{\text{MBP}}(\bm{x}) + V_{\text{FD}}(\bm{x}) \right) \\
f_4(\bm{x}) &= -\sum_{i=1}^{10} \max\left(g_i(\bm{x}), 0\right)
\end{align*}

\noindent
where the constraint functions \( g_i(\bm{x}) \) are defined as:
\begin{align*}
g_1(\bm{x}) &= 1 - 1.16 + 0.3717x_2x_4 + 0.0092928x_3 \\
g_2(\bm{x}) &= 0.32 - 0.261 + 0.0159x_1x_2 + 0.06486x_1 + 0.019x_2x_7 - 0.0144x_3x_5 - 0.0154464x_6 \\
g_3(\bm{x}) &= 0.32 - 0.214 - 0.00817x_5 + 0.045195x_1 + 0.0135168x_1 - 0.03099x_2x_6 \\
&\quad + 0.018x_2x_7 - 0.007176x_3 - 0.023223x_3 + 0.00364x_5x_6 + 0.018x_2^2 \\
g_4(\bm{x}) &= 0.32 - 0.74 + 0.61x_2 + 0.031296x_3 + 0.031872x_7 - 0.227x_2^2 \\
g_5(\bm{x}) &= 32 - 28.98 - 3.818x_3 + 4.2x_1x_2 - 1.27296x_6 + 2.68065x_7 \\
g_6(\bm{x}) &= 32 - 33.86 - 2.95x_3 + 5.057x_1x_2 + 3.795x_2 + 3.4431x_7 - 1.45728 \\
g_7(\bm{x}) &= 32 - 46.36 + 9.9x_2 + 4.4505x_1 \\
g_8(\bm{x}) &= 4 - f_2(\bm{x}) \\
g_9(\bm{x}) &= 9.9 - V_{\text{MBP}}(\bm{x}) \\
g_{10}(\bm{x}) &= 15.7 - V_{\text{FD}}(\bm{x})
\end{align*}

\noindent
with volume terms defined as:
\begin{align*}
V_{\text{MBP}}(\bm{x}) &= 10.58 - 0.674x_1x_2 - 0.67275x_2 \\
V_{\text{FD}}(\bm{x}) &= 16.45 - 0.489x_3x_7 - 0.8435x_6x_7.
\end{align*}

\noindent
The search space is defined as:
\[
\begin{aligned}
x_1 &\in [0.5, 1.5], \\
x_2 &\in [0.45, 1.35], \\
x_3, x_4 &\in [0.5, 1.5], \\
x_5 &\in [0.875, 2.625], \\
x_6, x_7 &\in [0.4, 1.2].
\end{aligned}
\]

\paragraph{Laser Plasma Acceleration.}

The laser plasma acceleration (LPA) problem considers the optimisation of beam quality in a laser-plasma accelerator~\citep{irshad2021expected,irshad2023multi}. 
It involves $m=3$ objectives with $d=4$ design variables. 
The objectives describe competing beam-quality metrics, such as improving the accelerated electron beam performance while controlling undesirable properties such as energy spread or beam instability. 
Following the practice in~\citep{ament2023unexpected,daulton2023hypervolume}, we use the public dataset from~\citep{irshad2023multi} to construct a surrogate model in our experiments. 

\clearpage
\newpage
\section{Additional Experimental Results}

\subsection{How Well Does CTD Perform Compared with  Well-Established Methods?}

\subsubsection{Noiseless Cases.}

\begin{table*}[!ht]
\centering
\caption{The HV results (mean and standard deviation) of the ten methods on the 20 benchmark and real-world problems under 200 evaluations. 
The method with the best mean HV is highlighted in bold. The symbols ``$-$'', ``$\sim$'' and ``$+$'' indicate that our method CTD is statistically worse than, equivalent to, and better than the competitor, respectively.}
\label{tab:noiseless_200}
\resizebox{\textwidth}{!}{
\begin{tabular}{l | ll| ll| ll| ll| ll| ll|ll}
\toprule
\bfseries Method & \multicolumn{2}{c|}{\bfseries DTLZ1 ($m=2$)} & \multicolumn{2}{c|}{\bfseries DTLZ2 ($m=2$)} & \multicolumn{2}{c|}{\bfseries DTLZ3 ($m=2$)} & \multicolumn{2}{c|}{\bfseries DTLZ4 ($m=2$)} & \multicolumn{2}{c|}{\bfseries DTLZ5 ($m=2$)} & \multicolumn{2}{c|}{\bfseries DTLZ6 ($m=2$)} & \multicolumn{2}{c}{\bfseries DTLZ7 ($m=2$)} \\
 & \multicolumn{1}{c}{Mean} & \multicolumn{1}{c|}{Std} & \multicolumn{1}{c}{Mean} & \multicolumn{1}{c|}{Std} & \multicolumn{1}{c}{Mean} & \multicolumn{1}{c|}{Std} & \multicolumn{1}{c}{Mean} & \multicolumn{1}{c|}{Std} & \multicolumn{1}{c}{Mean} & \multicolumn{1}{c|}{Std} & \multicolumn{1}{c}{Mean} & \multicolumn{1}{c|}{Std} & \multicolumn{1}{c}{Mean} & \multicolumn{1}{c}{Std} \\ \midrule
\bfseries Sobol & 5.13e-02 & 2.81e-01 ($\sim$) & 2.45e-02 & 2.33e-02 ($+$) & 3.26e-01 & 1.79e+00 ($\sim$) & 2.47e-02 & 2.69e-02 ($+$) & 1.76e-02 & 1.61e-02 ($+$) & 0.00e+00 & 0.00e+00 ($+$) & 0.00e+00 & 0.00e+00 ($+$) \\
\bfseries ParEGO & 2.56e+00 & 8.00e+00 ($\sim$) & 4.05e-01 & 2.23e-03 ($+$) & 3.91e+02 & 6.99e+02 ($\sim$) & 4.04e-01 & 2.04e-03 ($+$) & 4.03e-01 & 2.73e-03 ($+$) & 3.01e-01 & 1.13e-01 ($\sim$) & 6.32e-01 & 2.12e-01 ($-$) \\
\bfseries TS-TCH & 0.00e+00 & 0.00e+00 ($\sim$) & 1.49e-01 & 2.42e-02 ($+$) & 0.00e+00 & 0.00e+00 ($\sim$) & 1.51e-01 & 2.30e-02 ($+$) & 1.51e-01 & 2.23e-02 ($+$) & 0.00e+00 & 0.00e+00 ($+$) & 0.00e+00 & 0.00e+00 ($+$) \\
\bfseries PSL & \textbf{9.15e+01} & \textbf{1.44e+02} ($-$) & 3.94e-01 & 1.10e-02 ($+$) & 1.05e+03 & 1.62e+03 ($\sim$) & 1.17e-01 & 2.07e-02 ($+$) & 3.94e-01 & 1.10e-02 ($+$) & 0.00e+00 & 0.00e+00 ($+$) & 4.91e-01 & 1.63e-01 ($-$) \\
\bfseries JES & 0.00e+00 & 0.00e+00 ($\sim$) & 3.99e-01 & 2.77e-03 ($+$) & 9.96e+02 & 8.75e+02 ($\sim$) & 3.99e-01 & 3.57e-03 ($+$) & 4.00e-01 & 2.76e-03 ($+$) & 2.51e-01 & 7.30e-02 ($+$) & 6.45e-01 & 1.86e-01 ($-$) \\
\bfseries EHVI & 6.21e-01 & 3.40e+00 ($\sim$) & 4.08e-01 & 1.42e-03 ($+$) & 1.10e+03 & 1.35e+03 ($\sim$) & 4.08e-01 & 1.26e-03 ($+$) & 4.08e-01 & 1.75e-03 ($+$) & 2.63e-01 & 4.59e-02 ($+$) & 1.84e-01 & 3.21e-02 ($\sim$) \\
\bfseries MORBO & 0.00e+00 & 0.00e+00 ($\sim$) & 1.50e-01 & 2.21e-02 ($+$) & 0.00e+00 & 0.00e+00 ($\sim$) & 1.48e-01 & 1.94e-02 ($+$) & 1.66e-01 & 1.88e-02 ($+$) & 0.00e+00 & 0.00e+00 ($+$) & 0.00e+00 & 0.00e+00 ($+$) \\
\bfseries MOBO-OSD & 0.00e+00 & 0.00e+00 ($\sim$) & 3.73e-01 & 1.13e-02 ($+$) & \textbf{1.47e+03} & \textbf{9.24e+02} ($-$) & 2.51e-01 & 1.07e-01 ($+$) & 3.76e-01 & 8.43e-03 ($+$) & 2.03e-01 & 7.66e-02 ($+$) & \textbf{7.40e-01} & \textbf{1.35e-02} ($-$) \\
\bfseries qPOTS & 0.00e+00 & 0.00e+00 ($\sim$) & 3.37e-02 & 3.01e-02 ($+$) & 0.00e+00 & 0.00e+00 ($\sim$) & 3.62e-02 & 3.49e-02 ($+$) & 3.67e-02 & 3.93e-02 ($+$) & 0.00e+00 & 0.00e+00 ($+$) & 0.00e+00 & 0.00e+00 ($+$) \\
\bfseries CTD (ours) & 1.31e+01 & 3.48e+01 & \textbf{4.09e-01} & \textbf{1.35e-03} & 5.86e+02 & 9.79e+02 & \textbf{4.10e-01} & \textbf{1.24e-03} & \textbf{4.10e-01} & \textbf{1.11e-03} & \textbf{3.29e-01} & \textbf{3.49e-02} & 1.78e-01 & 3.00e-02 \\
\bottomrule
\toprule
\bfseries Method & \multicolumn{2}{c|}{\bfseries DTLZ1 ($m=3$)} & \multicolumn{2}{c|}{\bfseries DTLZ2 ($m=3$)} & \multicolumn{2}{c|}{\bfseries DTLZ3 ($m=3$)} & \multicolumn{2}{c|}{\bfseries DTLZ4 ($m=3$)} & \multicolumn{2}{c|}{\bfseries DTLZ5 ($m=3$)} & \multicolumn{2}{c|}{\bfseries DTLZ6 ($m=3$)} & \multicolumn{2}{c}{\bfseries DTLZ7 ($m=3$)} \\
 & \multicolumn{1}{c}{Mean} & \multicolumn{1}{c|}{Std} & \multicolumn{1}{c}{Mean} & \multicolumn{1}{c|}{Std} & \multicolumn{1}{c}{Mean} & \multicolumn{1}{c|}{Std} & \multicolumn{1}{c}{Mean} & \multicolumn{1}{c|}{Std} & \multicolumn{1}{c}{Mean} & \multicolumn{1}{c|}{Std} & \multicolumn{1}{c}{Mean} & \multicolumn{1}{c|}{Std} & \multicolumn{1}{c}{Mean} & \multicolumn{1}{c}{Std} \\ \midrule
\bfseries Sobol & 1.86e+03 & 4.49e+03 ($\sim$) & 5.43e-02 & 2.41e-02 ($+$) & 0.00e+00 & 0.00e+00 ($+$) & 5.71e-02 & 2.64e-02 ($+$) & 1.67e-03 & 2.00e-03 ($+$) & 0.00e+00 & 0.00e+00 ($+$) & 0.00e+00 & 0.00e+00 ($+$) \\
\bfseries ParEGO & 6.76e+03 & 9.88e+03 ($\sim$) & 5.86e-01 & 7.34e-02 ($+$) & 1.41e+06 & 9.03e+05 ($\sim$) & 5.75e-01 & 7.23e-02 ($+$) & 1.20e-01 & 3.74e-03 ($+$) & 9.69e-02 & 2.11e-02 ($+$) & 8.61e-01 & 1.29e-01 ($-$) \\
\bfseries TS-TCH & 9.99e+02 & 4.40e+03 ($+$) & 1.04e-01 & 4.29e-02 ($+$) & 0.00e+00 & 0.00e+00 ($+$) & 9.29e-02 & 4.47e-02 ($+$) & 2.07e-02 & 1.11e-02 ($+$) & 0.00e+00 & 0.00e+00 ($+$) & 0.00e+00 & 0.00e+00 ($+$) \\
\bfseries PSL & 1.01e+04 & 9.05e+03 ($\sim$) & 2.75e-01 & 1.23e-01 ($+$) & 2.41e+06 & 5.14e+05 ($\sim$) & 1.37e-01 & 5.12e-02 ($+$) & 1.26e-01 & 5.36e-03 ($\sim$) & 0.00e+00 & 0.00e+00 ($+$) & 2.64e-01 & 2.07e-01 ($-$) \\
\bfseries JES & 1.47e+04 & 4.10e+03 ($-$) & 5.46e-01 & 8.12e-02 ($+$) & 2.39e+06 & 6.57e+05 ($\sim$) & 5.62e-01 & 5.97e-02 ($+$) & 1.15e-01 & 6.52e-03 ($+$) & 7.35e-02 & 2.19e-02 ($+$) & 7.71e-01 & 1.91e-01 ($-$) \\
\bfseries EHVI & 1.39e+04 & 6.10e+03 ($-$) & 3.23e-01 & 1.20e-01 ($+$) & 2.45e+06 & 6.89e+05 ($\sim$) & 3.31e-01 & 1.12e-01 ($+$) & 1.11e-01 & 1.79e-02 ($+$) & 7.86e-02 & 1.49e-02 ($+$) & 2.27e-01 & 7.35e-03 ($+$) \\
\bfseries MORBO & 2.35e+03 & 6.92e+03 ($\sim$) & 1.14e-01 & 5.52e-02 ($+$) & 0.00e+00 & 0.00e+00 ($+$) & 1.02e-01 & 5.37e-02 ($+$) & 1.65e-02 & 8.94e-03 ($+$) & 0.00e+00 & 0.00e+00 ($+$) & 0.00e+00 & 0.00e+00 ($+$) \\
\bfseries MOBO-OSD & \textbf{1.66e+04} & \textbf{3.47e+03} ($-$) & 2.51e-01 & 7.23e-02 ($+$) & \textbf{3.63e+06} & \textbf{6.96e+05} ($-$) & 2.58e-02 & 8.77e-02 ($+$) & 6.60e-02 & 2.11e-02 ($+$) & 3.71e-02 & 2.67e-02 ($+$) & \textbf{9.31e-01} & \textbf{2.49e-02} ($-$) \\
\bfseries qPOTS & 0.00e+00 & 0.00e+00 ($+$) & 0.00e+00 & 0.00e+00 ($+$) & 0.00e+00 & 0.00e+00 ($+$) & 0.00e+00 & 0.00e+00 ($+$) & 0.00e+00 & 0.00e+00 ($+$) & 0.00e+00 & 0.00e+00 ($+$) & 0.00e+00 & 0.00e+00 ($+$) \\
\bfseries CTD (ours) & 6.15e+03 & 7.35e+03 & \textbf{6.43e-01} & \textbf{3.14e-02} & 1.91e+06 & 1.00e+06 & \textbf{6.45e-01} & \textbf{2.70e-02} & \textbf{1.29e-01} & \textbf{2.18e-03} & \textbf{1.07e-01} & \textbf{1.57e-02} & 2.28e-01 & 6.42e-04 \\
\bottomrule
\toprule
\bfseries Method & \multicolumn{2}{c|}{\bfseries Four bar truss design} & \multicolumn{2}{c|}{\bfseries Pressure vessel design} & \multicolumn{2}{c|}{\bfseries Hatch cover design} & \multicolumn{2}{c|}{\bfseries Vehicle safety} & \multicolumn{2}{c|}{\bfseries Car side impact} & \multicolumn{2}{c|}{\bfseries LPA} & \multicolumn{2}{c}{\bfseries Summary} \\
 & \multicolumn{1}{c}{Mean} & \multicolumn{1}{c|}{Std} & \multicolumn{1}{c}{Mean} & \multicolumn{1}{c|}{Std} & \multicolumn{1}{c}{Mean} & \multicolumn{1}{c|}{Std} & \multicolumn{1}{c}{Mean} & \multicolumn{1}{c|}{Std} & \multicolumn{1}{c}{Mean} & \multicolumn{1}{c|}{Std} & \multicolumn{1}{c}{Mean} & \multicolumn{1}{c|}{Std} & \multicolumn{2}{c}{$-$/$\sim$/$+$} \\ \midrule
\bfseries Sobol & 4.65e+01 & 5.45e-01 ($+$) & 5.45e+09 & 1.70e+09 ($+$) & 2.10e+04 & 2.38e+02 ($+$) & 1.61e+01 & 8.15e-01 ($+$) & 2.42e+02 & 3.96e+00 ($+$) & 3.08e+07 & 2.43e+06 ($+$) & \multicolumn{2}{c}{0/3/17} \\
\bfseries ParEGO & 5.34e+01 & 9.61e-01 ($+$) & 1.11e+10 & 2.86e+07 ($-$) & 2.18e+04 & 1.09e+01 ($+$) & 2.68e+01 & 5.05e-01 ($-$) & 3.15e+02 & 1.07e+02 ($+$) & 4.71e+07 & 4.94e+06 ($+$) & \multicolumn{2}{c}{4/5/11} \\
\bfseries TS-TCH & 5.20e+01 & 8.68e-01 ($+$) & 1.05e+10 & 2.11e+08 ($+$) & 2.17e+04 & 5.49e+01 ($+$) & 2.35e+01 & 4.11e-01 ($+$) & 2.71e+02 & 9.21e+01 ($+$) & 4.38e+07 & 2.96e+06 ($+$) & \multicolumn{2}{c}{0/2/18} \\
\bfseries PSL & 5.40e+01 & 3.69e-02 ($+$) & 1.09e+10 & 1.67e+08 ($\sim$) & 2.18e+04 & 7.26e+00 ($+$) & 2.38e+01 & 1.48e+00 ($+$) & 3.31e+02 & 5.83e+00 ($\sim$) & 5.00e+07 & 2.91e+06 ($+$) & \multicolumn{2}{c}{3/6/11} \\
\bfseries JES & 5.34e+01 & 1.01e+00 ($+$) & 1.09e+10 & 8.06e+07 ($+$) & 2.18e+04 & 2.92e+01 ($+$) & \textbf{2.71e+01} & \textbf{3.63e-01} ($-$) & \textbf{3.46e+02} & \textbf{8.98e+00} ($-$) & 4.69e+07 & 4.69e+06 ($+$) & \multicolumn{2}{c}{5/3/12} \\
\bfseries EHVI & 5.43e+01 & 7.44e-02 ($+$) & 1.10e+10 & 9.64e+07 ($\sim$) & \textbf{2.18e+04} & \textbf{7.04e+00} ($\sim$) & 2.59e+01 & 7.30e-01 ($+$) & 2.94e+02 & 1.00e+02 ($\sim$) & \textbf{5.86e+07} & \textbf{2.70e+06} ($\sim$) & \multicolumn{2}{c}{1/8/11} \\
\bfseries MORBO & 5.21e+01 & 7.71e-01 ($+$) & 1.05e+10 & 1.99e+08 ($+$) & 2.17e+04 & 6.28e+01 ($+$) & 2.33e+01 & 4.64e-01 ($+$) & 3.01e+02 & 3.37e+00 ($+$) & 4.39e+07 & 2.63e+06 ($+$) & \multicolumn{2}{c}{0/3/17} \\
\bfseries MOBO-OSD & 5.21e+01 & 8.54e-01 ($+$) & \textbf{1.11e+10} & \textbf{1.40e+07} ($-$) & 2.15e+04 & 2.76e+02 ($+$) & 2.38e+01 & 1.10e+00 ($+$) & 2.84e+02 & 1.11e+01 ($+$) & 4.07e+07 & 3.76e+06 ($+$) & \multicolumn{2}{c}{6/1/13} \\
\bfseries qPOTS & 3.98e+01 & 1.31e+00 ($+$) & 1.10e+10 & 4.88e+07 ($\sim$) & 1.88e+04 & 4.14e+03 ($+$) & 1.32e+01 & 2.38e+00 ($+$) & 2.53e+02 & 1.01e+01 ($+$) & 4.38e+07 & 2.55e+06 ($+$) & \multicolumn{2}{c}{0/3/17} \\
\bfseries CTD (ours) & \textbf{5.43e+01} & \textbf{2.80e-02} & 1.09e+10 & 1.91e+08 & 2.18e+04 & 1.52e+01 & 2.66e+01 & 6.20e-01 & 3.26e+02 & 1.07e+01 & 5.82e+07 & 3.97e+06 & \multicolumn{2}{c}{--/--/--} \\
\bottomrule
\end{tabular}
}
\end{table*}

\clearpage
\subsubsection{Noisy Cases.}\label{appendix:sec:noisy}

\begin{table*}[!ht]
\centering
\caption{The HV results (mean and standard deviation) of the eight methods on the 20 benchmark and real-world problems under noisy settings and 200 evaluations. 
The method with the best mean HV is highlighted in bold. The symbols ``$-$'', ``$\sim$'' and ``$+$'' indicate that the proposed CTD is statistically worse than, equivalent to, and better than the competitor, respectively.}
\label{tab:noisy_200}
\resizebox{\textwidth}{!}{
\begin{tabular}{l | ll| ll| ll| ll| ll| ll|ll}
\toprule
\bfseries Method & \multicolumn{2}{c|}{\bfseries DTLZ1 ($m=2$)} & \multicolumn{2}{c|}{\bfseries DTLZ2 ($m=2$)} & \multicolumn{2}{c|}{\bfseries DTLZ3 ($m=2$)} & \multicolumn{2}{c|}{\bfseries DTLZ4 ($m=2$)} & \multicolumn{2}{c|}{\bfseries DTLZ5 ($m=2$)} & \multicolumn{2}{c|}{\bfseries DTLZ6 ($m=2$)} & \multicolumn{2}{c}{\bfseries DTLZ7 ($m=2$)} \\
 & \multicolumn{1}{c}{Mean} & \multicolumn{1}{c|}{Std} & \multicolumn{1}{c}{Mean} & \multicolumn{1}{c|}{Std} & \multicolumn{1}{c}{Mean} & \multicolumn{1}{c|}{Std} & \multicolumn{1}{c}{Mean} & \multicolumn{1}{c|}{Std} & \multicolumn{1}{c}{Mean} & \multicolumn{1}{c|}{Std} & \multicolumn{1}{c}{Mean} & \multicolumn{1}{c|}{Std} & \multicolumn{1}{c}{Mean} & \multicolumn{1}{c}{Std} \\ \midrule
\bfseries Sobol & 3.20e-01 & 1.75e+00 ($\sim$) & 2.08e-02 & 1.60e-02 ($+$) & 1.10e+02 & 6.03e+02 ($+$) & 2.34e-02 & 2.02e-02 ($+$) & 2.18e-02 & 2.17e-02 ($+$) & 0.00e+00 & 0.00e+00 ($+$) & 0.00e+00 & 0.00e+00 ($\sim$) \\
\bfseries ParEGO & 0.00e+00 & 0.00e+00 ($\sim$) & 1.42e-01 & 4.65e-02 ($\sim$) & 6.00e+02 & 7.25e+02 ($\sim$) & \textbf{1.39e-01} & \textbf{7.34e-02} ($\sim$) & \textbf{1.40e-01} & \textbf{8.26e-02} ($\sim$) & \textbf{2.27e-01} & \textbf{7.34e-02} ($\sim$) & \textbf{1.04e-01} & \textbf{1.55e-01} ($\sim$) \\
\bfseries TS-TCH & 0.00e+00 & 0.00e+00 ($\sim$) & 5.39e-03 & 1.06e-02 ($+$) & 0.00e+00 & 0.00e+00 ($+$) & 8.57e-03 & 1.40e-02 ($+$) & 1.35e-02 & 2.17e-02 ($+$) & 0.00e+00 & 0.00e+00 ($+$) & 0.00e+00 & 0.00e+00 ($\sim$) \\
\bfseries JES & 0.00e+00 & 0.00e+00 ($\sim$) & \textbf{1.43e-01} & \textbf{6.86e-02} ($\sim$) & 8.24e+02 & 9.69e+02 ($\sim$) & 1.06e-01 & 5.03e-02 ($\sim$) & 9.97e-02 & 6.62e-02 ($\sim$) & 2.17e-01 & 6.15e-02 ($\sim$) & 1.37e-02 & 3.91e-02 ($\sim$) \\
\bfseries EHVI & 0.00e+00 & 0.00e+00 ($\sim$) & 6.86e-02 & 8.47e-02 ($+$) & \textbf{8.92e+02} & \textbf{8.95e+02} ($\sim$) & 5.31e-02 & 5.55e-02 ($+$) & 8.97e-02 & 7.69e-02 ($\sim$) & 1.94e-01 & 5.54e-02 ($\sim$) & 8.76e-02 & 1.61e-01 ($\sim$) \\
\bfseries MORBO & 2.88e-01 & 1.58e+00 ($\sim$) & 4.13e-03 & 8.07e-03 ($+$) & 0.00e+00 & 0.00e+00 ($+$) & 5.99e-03 & 1.05e-02 ($+$) & 7.80e-03 & 1.23e-02 ($+$) & 0.00e+00 & 0.00e+00 ($+$) & 0.00e+00 & 0.00e+00 ($\sim$) \\
\bfseries qPOTS & \textbf{1.88e+00} & \textbf{1.03e+01} ($\sim$) & 7.95e-03 & 1.17e-02 ($+$) & 0.00e+00 & 0.00e+00 ($+$) & 7.90e-03 & 1.63e-02 ($+$) & 8.16e-03 & 1.72e-02 ($+$) & 0.00e+00 & 0.00e+00 ($+$) & 0.00e+00 & 0.00e+00 ($\sim$) \\
\bfseries CTD (ours) & 8.72e-03 & 4.78e-02 & 1.37e-01 & 6.87e-02 & 6.03e+02 & 8.58e+02 & 1.33e-01 & 4.83e-02 & 1.33e-01 & 4.38e-02 & 2.07e-01 & 8.12e-02 & 3.02e-02 & 1.06e-01 \\
\bottomrule
\toprule
\bfseries Method & \multicolumn{2}{c|}{\bfseries DTLZ1 ($m=3$)} & \multicolumn{2}{c|}{\bfseries DTLZ2 ($m=3$)} & \multicolumn{2}{c|}{\bfseries DTLZ3 ($m=3$)} & \multicolumn{2}{c|}{\bfseries DTLZ4 ($m=3$)} & \multicolumn{2}{c|}{\bfseries DTLZ5 ($m=3$)} & \multicolumn{2}{c|}{\bfseries DTLZ6 ($m=3$)} & \multicolumn{2}{c}{\bfseries DTLZ7 ($m=3$)} \\
 & \multicolumn{1}{c}{Mean} & \multicolumn{1}{c|}{Std} & \multicolumn{1}{c}{Mean} & \multicolumn{1}{c|}{Std} & \multicolumn{1}{c}{Mean} & \multicolumn{1}{c|}{Std} & \multicolumn{1}{c}{Mean} & \multicolumn{1}{c|}{Std} & \multicolumn{1}{c}{Mean} & \multicolumn{1}{c|}{Std} & \multicolumn{1}{c}{Mean} & \multicolumn{1}{c|}{Std} & \multicolumn{1}{c}{Mean} & \multicolumn{1}{c}{Std} \\ \midrule
\bfseries Sobol & 2.20e+03 & 4.59e+03 ($+$) & 5.55e-02 & 2.42e-02 ($+$) & 0.00e+00 & 0.00e+00 ($+$) & 5.90e-02 & 2.20e-02 ($+$) & 2.07e-03 & 3.37e-03 ($+$) & 0.00e+00 & 0.00e+00 ($+$) & 0.00e+00 & 0.00e+00 ($\sim$) \\
\bfseries ParEGO & 1.04e+04 & 5.79e+03 ($\sim$) & 1.49e-01 & 8.05e-02 ($+$) & 1.86e+06 & 6.94e+05 ($\sim$) & 1.39e-01 & 9.21e-02 ($+$) & 2.20e-02 & 1.66e-02 ($+$) & 5.81e-02 & 2.93e-02 ($\sim$) & \textbf{4.90e-02} & \textbf{1.48e-01} ($\sim$) \\
\bfseries TS-TCH & 1.59e+03 & 2.67e+03 ($+$) & 1.43e-02 & 1.59e-02 ($+$) & 0.00e+00 & 0.00e+00 ($+$) & 1.78e-02 & 1.71e-02 ($+$) & 6.97e-04 & 2.19e-03 ($+$) & 0.00e+00 & 0.00e+00 ($+$) & 0.00e+00 & 0.00e+00 ($\sim$) \\
\bfseries JES & \textbf{1.38e+04} & \textbf{4.54e+03} ($\sim$) & 1.12e-01 & 8.10e-02 ($+$) & 1.91e+06 & 8.83e+05 ($\sim$) & 1.25e-01 & 7.43e-02 ($+$) & 1.54e-02 & 1.27e-02 ($+$) & 5.22e-02 & 2.97e-02 ($\sim$) & 2.13e-02 & 7.27e-02 ($\sim$) \\
\bfseries EHVI & 1.37e+04 & 5.96e+03 ($\sim$) & 2.18e-02 & 4.04e-02 ($+$) & \textbf{2.44e+06} & \textbf{5.12e+05} ($\sim$) & 2.52e-02 & 3.46e-02 ($+$) & 3.45e-03 & 7.29e-03 ($+$) & 6.52e-02 & 2.47e-02 ($\sim$) & 4.62e-02 & 1.07e-01 ($\sim$) \\
\bfseries MORBO & 2.12e+03 & 5.59e+03 ($+$) & 1.24e-02 & 1.13e-02 ($+$) & 0.00e+00 & 0.00e+00 ($+$) & 2.05e-02 & 2.79e-02 ($+$) & 3.30e-04 & 7.80e-04 ($+$) & 0.00e+00 & 0.00e+00 ($+$) & 0.00e+00 & 0.00e+00 ($\sim$) \\
\bfseries qPOTS & 2.62e+03 & 6.00e+03 ($+$) & 1.56e-02 & 1.81e-02 ($+$) & 2.81e+04 & 1.54e+05 ($+$) & 1.32e-02 & 1.62e-02 ($+$) & 3.61e-04 & 9.12e-04 ($+$) & 0.00e+00 & 0.00e+00 ($+$) & 0.00e+00 & 0.00e+00 ($\sim$) \\
\bfseries CTD (ours) & 1.07e+04 & 7.98e+03 & \textbf{1.98e-01} & \textbf{6.82e-02} & 2.15e+06 & 4.99e+05 & \textbf{1.98e-01} & \textbf{6.62e-02} & \textbf{3.28e-02} & \textbf{1.38e-02} & \textbf{6.82e-02} & \textbf{2.66e-02} & 4.97e-03 & 2.72e-02 \\
\bottomrule
\toprule
\bfseries Method & \multicolumn{2}{c|}{\bfseries Four bar truss design} & \multicolumn{2}{c|}{\bfseries Pressure vessel design} & \multicolumn{2}{c|}{\bfseries Hatch cover design} & \multicolumn{2}{c|}{\bfseries Vehicle safety} & \multicolumn{2}{c|}{\bfseries Car side impact} & \multicolumn{2}{c|}{\bfseries LPA} & \multicolumn{2}{c}{\bfseries Summary} \\
 & \multicolumn{1}{c}{Mean} & \multicolumn{1}{c|}{Std} & \multicolumn{1}{c}{Mean} & \multicolumn{1}{c|}{Std} & \multicolumn{1}{c}{Mean} & \multicolumn{1}{c|}{Std} & \multicolumn{1}{c}{Mean} & \multicolumn{1}{c|}{Std} & \multicolumn{1}{c}{Mean} & \multicolumn{1}{c|}{Std} & \multicolumn{1}{c}{Mean} & \multicolumn{1}{c|}{Std} & \multicolumn{2}{c}{$-$/$\sim$/$+$} \\ \midrule
\bfseries Sobol & 4.65e+01 & 6.07e-01 ($+$) & 5.28e+09 & 1.70e+09 ($+$) & 2.05e+04 & 4.43e+02 ($+$) & 1.62e+01 & 6.87e-01 ($+$) & 2.42e+02 & 5.97e+00 ($+$) & 2.90e+07 & 2.82e+06 ($+$) & \multicolumn{2}{c}{0/3/17} \\
\bfseries ParEGO & 5.23e+01 & 1.13e+00 ($\sim$) & 1.04e+10 & 4.70e+08 ($+$) & 2.16e+04 & 8.50e+01 ($\sim$) & 2.67e+01 & 3.60e-01 ($\sim$) & 2.63e+02 & 1.55e+01 ($+$) & 4.77e+07 & 4.94e+06 ($+$) & \multicolumn{2}{c}{0/14/6} \\
\bfseries TS-TCH & 5.08e+01 & 8.96e-01 ($+$) & 9.00e+09 & 1.01e+09 ($+$) & 2.14e+04 & 2.45e+02 ($+$) & 2.26e+01 & 5.55e-01 ($+$) & 2.60e+02 & 1.13e+01 ($+$) & 4.33e+07 & 2.52e+06 ($+$) & \multicolumn{2}{c}{0/3/17} \\
\bfseries JES & 4.61e+01 & 2.13e+00 ($+$) & 1.06e+10 & 2.84e+08 ($\sim$) & 2.12e+04 & 5.83e+02 ($+$) & 2.36e+01 & 1.49e+00 ($+$) & 2.74e+02 & 1.84e+01 ($+$) & 4.65e+07 & 4.95e+06 ($+$) & \multicolumn{2}{c}{0/12/8} \\
\bfseries EHVI & 5.18e+01 & 1.86e+00 ($\sim$) & \textbf{1.07e+10} & \textbf{2.96e+08} ($\sim$) & 2.15e+04 & 3.72e+02 ($\sim$) & \textbf{2.72e+01} & \textbf{3.44e-01} ($-$) & \textbf{3.28e+02} & \textbf{1.11e+01} ($-$) & 5.78e+07 & 3.75e+06 ($\sim$) & \multicolumn{2}{c}{2/13/5} \\
\bfseries MORBO & 5.04e+01 & 1.26e+00 ($+$) & 8.99e+09 & 9.19e+08 ($+$) & 2.15e+04 & 3.67e+02 ($+$) & 2.24e+01 & 6.05e-01 ($+$) & 2.59e+02 & 1.04e+01 ($+$) & 4.37e+07 & 2.98e+06 ($+$) & \multicolumn{2}{c}{0/3/17} \\
\bfseries qPOTS & 5.05e+01 & 9.04e-01 ($+$) & 9.28e+09 & 8.85e+08 ($+$) & 2.15e+04 & 2.09e+02 ($+$) & 2.25e+01 & 5.79e-01 ($+$) & 2.63e+02 & 9.98e+00 ($+$) & 4.38e+07 & 3.23e+06 ($+$) & \multicolumn{2}{c}{0/3/17} \\
\bfseries CTD (ours) & \textbf{5.26e+01} & \textbf{7.99e-01} & 1.07e+10 & 2.52e+08 & \textbf{2.17e+04} & \textbf{7.08e+01} & 2.66e+01 & 3.16e-01 & 2.93e+02 & 1.59e+01 & \textbf{5.80e+07} & \textbf{4.10e+06} & \multicolumn{2}{c}{--/--/--} \\
\bottomrule
\end{tabular}
}
\end{table*}



\subsubsection{High-Dimensional Problems.}\label{appendix:sec:HD}

\begin{table*}[!ht]
\centering
\caption{The HV results (mean and standard deviation) of the ten methods on 14 problems with $d=50$. 
The method with the best mean HV is highlighted in bold. The symbols ``$-$'', ``$\sim$'' and ``$+$'' indicate that the proposed CTD is statistically worse than, equivalent to, and better than the competitor, respectively.}
\label{tab:hd}
\resizebox{\textwidth}{!}{
\begin{tabular}{l | ll| ll| ll| ll| ll| ll|ll}
\toprule
\bfseries Method & \multicolumn{2}{c|}{\bfseries DTLZ1 ($m=2$)} & \multicolumn{2}{c|}{\bfseries DTLZ2 ($m=2$)} & \multicolumn{2}{c|}{\bfseries DTLZ3 ($m=2$)} & \multicolumn{2}{c|}{\bfseries DTLZ4 ($m=2$)} & \multicolumn{2}{c|}{\bfseries DTLZ5 ($m=2$)} & \multicolumn{2}{c|}{\bfseries DTLZ6 ($m=2$)} & \multicolumn{2}{c}{\bfseries DTLZ7 ($m=2$)} \\
 & \multicolumn{1}{c}{Mean} & \multicolumn{1}{c|}{Std} & \multicolumn{1}{c}{Mean} & \multicolumn{1}{c|}{Std} & \multicolumn{1}{c}{Mean} & \multicolumn{1}{c|}{Std} & \multicolumn{1}{c}{Mean} & \multicolumn{1}{c|}{Std} & \multicolumn{1}{c}{Mean} & \multicolumn{1}{c|}{Std} & \multicolumn{1}{c}{Mean} & \multicolumn{1}{c|}{Std} & \multicolumn{1}{c}{Mean} & \multicolumn{1}{c}{Std} \\ \midrule
\bfseries Sobol & 0.00e+00 & 0.00e+00 ($+$) & 0.00e+00 & 0.00e+00 ($+$) & 0.00e+00 & 0.00e+00 ($+$) & 0.00e+00 & 0.00e+00 ($+$) & 0.00e+00 & 0.00e+00 ($+$) & 0.00e+00 & 0.00e+00 ($+$) & 0.00e+00 & 0.00e+00 ($+$)\\
\bfseries ParEGO & 1.34e+05 & 2.41e+04 ($+$) & 6.48e-02 & 9.40e-02 ($+$) & 1.91e+05 & 5.93e+04 ($+$) & 2.08e-02 & 4.07e-02 ($+$) & 2.92e-01 & 3.36e-01 ($+$) & 8.85e-02 & 5.30e-02 ($+$) & 4.01e-01 & 2.11e-01 ($-$)\\
\bfseries TS-TCH & 0.00e+00 & 0.00e+00 ($+$) & 0.00e+00 & 0.00e+00 ($+$) & 0.00e+00 & 0.00e+00 ($+$) & 0.00e+00 & 0.00e+00 ($+$) & 0.00e+00 & 0.00e+00 ($+$) & 0.00e+00 & 0.00e+00 ($+$) & 0.00e+00 & 0.00e+00 ($+$)\\
\bfseries PSL & 8.86e+04 & 2.98e+04 ($+$) & 0.00e+00 & 0.00e+00 ($+$) & 3.99e+04 & 4.16e+04 ($+$) & 0.00e+00 & 0.00e+00 ($+$) & 0.00e+00 & 0.00e+00 ($+$) & 0.00e+00 & 0.00e+00 ($+$) & 3.50e-01 & 2.02e-01 ($\sim$)\\
\bfseries JES & 0.00e+00 & 0.00e+00 ($+$) & 0.00e+00 & 0.00e+00 ($+$) & 0.00e+00 & 0.00e+00 ($+$) & 0.00e+00 & 0.00e+00 ($+$) & 0.00e+00 & 0.00e+00 ($+$) & 0.00e+00 & 0.00e+00 ($+$) & 0.00e+00 & 0.00e+00 ($+$)\\
\bfseries EHVI & \textbf{2.04e+05} & \textbf{1.48e+04} ($\sim$) & 0.00e+00 & 0.00e+00 ($+$) & 2.78e+05 & 7.70e+04 ($\sim$) & 0.00e+00 & 0.00e+00 ($+$) & 0.00e+00 & 0.00e+00 ($+$) & 1.05e-01 & 6.48e-02 ($+$) & 2.10e-01 & 2.64e-02 ($\sim$)\\
\bfseries MORBO & 0.00e+00 & 0.00e+00 ($+$) & 0.00e+00 & 0.00e+00 ($+$) & 0.00e+00 & 0.00e+00 ($+$) & 0.00e+00 & 0.00e+00 ($+$) & 0.00e+00 & 0.00e+00 ($+$) & 0.00e+00 & 0.00e+00 ($+$) & 0.00e+00 & 0.00e+00 ($+$)\\
\bfseries MOBO-OSD & 1.77e+05 & 2.29e+04 ($+$) & 0.00e+00 & 0.00e+00 ($+$) & \textbf{3.19e+05} & \textbf{6.53e+04} ($\sim$) & 3.80e-02 & 7.14e-02 ($+$) & 9.29e-02 & 1.97e-01 ($+$) & 2.00e-03 & 1.10e-02 ($+$) & \textbf{7.39e-01} & \textbf{1.06e-02} ($-$)\\
\bfseries qPOTS & 0.00e+00 & 0.00e+00 ($+$) & 0.00e+00 & 0.00e+00 ($+$) & 0.00e+00 & 0.00e+00 ($+$) & 0.00e+00 & 0.00e+00 ($+$) & 0.00e+00 & 0.00e+00 ($+$) & 0.00e+00 & 0.00e+00 ($+$) & 0.00e+00 & 0.00e+00 ($+$)\\
\bfseries CTD (ours) & 1.94e+05 & 2.83e+04 & \textbf{1.84e-01} & \textbf{1.59e-01} & 2.97e+05 & 7.55e+04 & \textbf{1.29e-01} & \textbf{1.17e-01} & \textbf{7.42e-01} & \textbf{4.11e-01} & \textbf{1.56e-01} & \textbf{9.46e-02} & 2.11e-01 & 2.45e-02\\
\bottomrule
\toprule
\bfseries Method & \multicolumn{2}{c|}{\bfseries DTLZ1 ($m=3$)} & \multicolumn{2}{c|}{\bfseries DTLZ2 ($m=3$)} & \multicolumn{2}{c|}{\bfseries DTLZ3 ($m=3$)} & \multicolumn{2}{c|}{\bfseries DTLZ4 ($m=3$)} & \multicolumn{2}{c|}{\bfseries DTLZ5 ($m=3$)} & \multicolumn{2}{c|}{\bfseries DTLZ6 ($m=3$)} & \multicolumn{2}{c}{\bfseries DTLZ7 ($m=3$)} \\
 & \multicolumn{1}{c}{Mean} & \multicolumn{1}{c|}{Std} & \multicolumn{1}{c}{Mean} & \multicolumn{1}{c|}{Std} & \multicolumn{1}{c}{Mean} & \multicolumn{1}{c|}{Std} & \multicolumn{1}{c}{Mean} & \multicolumn{1}{c|}{Std} & \multicolumn{1}{c}{Mean} & \multicolumn{1}{c|}{Std} & \multicolumn{1}{c}{Mean} & \multicolumn{1}{c|}{Std} & \multicolumn{1}{c}{Mean} & \multicolumn{1}{c}{Std} \\ \midrule
\bfseries Sobol & 0.00e+00 & 0.00e+00 ($+$) & 2.27e+00 & 7.62e-01 ($+$) & 0.00e+00 & 0.00e+00 ($+$) & 1.67e-01 & 2.06e-01 ($+$) & 9.90e-01 & 5.03e-01 ($+$) & 0.00e+00 & 0.00e+00 ($+$) & 0.00e+00 & 0.00e+00 ($+$)\\
\bfseries ParEGO & 9.98e+07 & 1.71e+07 ($+$) & 3.19e+00 & 2.38e+00 ($+$) & 4.21e+08 & 1.17e+08 ($+$) & 7.42e-01 & 1.28e+00 ($+$) & 4.42e+00 & 4.48e+00 ($+$) & 1.83e-02 & 2.85e-02 ($+$) & 6.84e-01 & 2.18e-01 ($-$)\\
\bfseries TS-TCH & 0.00e+00 & 0.00e+00 ($+$) & 1.66e+00 & 6.99e-01 ($+$) & 0.00e+00 & 0.00e+00 ($+$) & 1.64e-01 & 1.93e-01 ($+$) & 6.29e-01 & 3.69e-01 ($+$) & 0.00e+00 & 0.00e+00 ($+$) & 0.00e+00 & 0.00e+00 ($+$)\\
\bfseries PSL & 8.34e+07 & 2.16e+07 ($+$) & 3.56e+00 & 1.83e-02 ($+$) & 2.47e+08 & 1.36e+08 ($+$) & 3.85e-01 & 8.41e-01 ($+$) & 1.15e+00 & 0.00e+00 ($+$) & 0.00e+00 & 0.00e+00 ($+$) & 4.66e-01 & 1.78e-01 ($-$)\\
\bfseries JES & 0.00e+00 & 0.00e+00 ($+$) & 1.61e+00 & 8.66e-01 ($+$) & 0.00e+00 & 0.00e+00 ($+$) & 2.26e-01 & 2.79e-01 ($+$) & 6.22e-01 & 4.00e-01 ($+$) & 0.00e+00 & 0.00e+00 ($+$) & 0.00e+00 & 0.00e+00 ($+$)\\
\bfseries EHVI & 1.81e+08 & 1.40e+07 ($\sim$) & 1.98e+00 & 1.35e+00 ($+$) & 5.33e+08 & 1.00e+08 ($\sim$) & 2.06e-01 & 3.63e-01 ($+$) & 8.15e-01 & 8.83e-01 ($+$) & 5.04e-02 & 2.29e-02 ($+$) & 2.64e-01 & 1.21e-01 ($\sim$)\\
\bfseries MORBO & 0.00e+00 & 0.00e+00 ($+$) & 1.75e+00 & 9.17e-01 ($+$) & 0.00e+00 & 0.00e+00 ($+$) & 1.81e-01 & 4.48e-01 ($+$) & 6.41e-01 & 4.20e-01 ($+$) & 0.00e+00 & 0.00e+00 ($+$) & 0.00e+00 & 0.00e+00 ($+$)\\
\bfseries MOBO-OSD & \textbf{1.88e+08} & \textbf{4.93e+06} ($-$) & 1.83e+00 & 1.29e+00 ($+$) & \textbf{7.87e+08} & \textbf{4.76e+07} ($-$) & 3.16e+00 & 3.35e+00 ($+$) & 1.49e+00 & 2.71e+00 ($+$) & 3.02e-03 & 1.06e-02 ($+$) & \textbf{8.72e-01} & \textbf{7.41e-02} ($-$)\\
\bfseries qPOTS & 0.00e+00 & 0.00e+00 ($+$) & 2.20e+00 & 4.52e-16 ($+$) & 0.00e+00 & 0.00e+00 ($+$) & 0.00e+00 & 0.00e+00 ($+$) & 1.04e+00 & 0.00e+00 ($+$) & 0.00e+00 & 0.00e+00 ($+$) & 0.00e+00 & 0.00e+00 ($+$)\\
\bfseries CTD (ours) & 1.80e+08 & 1.11e+07 & \textbf{1.44e+01} & \textbf{2.97e+00} & 5.09e+08 & 9.43e+07 & \textbf{6.47e+00} & \textbf{2.16e+00} & \textbf{1.49e+01} & \textbf{4.24e+00} & \textbf{6.41e-02} & \textbf{1.56e-02} & 2.50e-01 & 6.98e-02\\
\bottomrule
\end{tabular}
}
\end{table*}

\begin{table*}[!ht]
\centering
\caption{The summary of statistical results for the noisy experiments in Table~\ref{tab:hd}. Here, the left, median, and right numbers are the counts of test problems where the CTD was statistically worse, equivalent, or better to the peer method, respectively.}
\label{tab:hd_summary}
\resizebox{\textwidth}{!}{
\begin{tabular}{lccccccccc}
\toprule
 & \bfseries Sobol & \bfseries ParEGO & \bfseries TS-TCH & \bfseries PSL & \bfseries JES & \bfseries EHVI & \bfseries MORBO & \bfseries MOBO-OSD & \bfseries qPOTS \\
\midrule
\bfseries CTD (ours) & 0/0/14 & 2/0/12 & 0/0/14 & 1/1/12 & 0/0/14 & 0/6/8 & 0/0/14 & 4/1/9 & 0/0/14 \\
\bottomrule
\end{tabular}
}
\end{table*}

\clearpage
\subsection{Sensitivity Analysis}\label{appendix:sec:sensitivity}

\begin{table*}[!ht]
\centering
\caption{The HV results (mean and standard deviation) of our method with six different threshold on the 20 benchmark and real-world problems under 200 evaluations. 
The method with the best mean HV is highlighted in bold. The symbols ``$-$'', ``$\sim$'' and ``$+$'' indicate that our method CTD is statistically worse than, equivalent to, and better than the competitor, respectively.}
\label{tab:placeholder}
\resizebox{\textwidth}{!}{
\begin{tabular}{l | ll| ll| ll| ll| ll| ll|ll}
\toprule
\bfseries Method & \multicolumn{2}{c|}{\bfseries DTLZ1 ($m=2$)} & \multicolumn{2}{c|}{\bfseries DTLZ2 ($m=2$)} & \multicolumn{2}{c|}{\bfseries DTLZ3 ($m=2$)} & \multicolumn{2}{c|}{\bfseries DTLZ4 ($m=2$)} & \multicolumn{2}{c|}{\bfseries DTLZ5 ($m=2$)} & \multicolumn{2}{c|}{\bfseries DTLZ6 ($m=2$)} & \multicolumn{2}{c}{\bfseries DTLZ7 ($m=2$)} \\
 & \multicolumn{1}{c}{Mean} & \multicolumn{1}{c|}{Std} & \multicolumn{1}{c}{Mean} & \multicolumn{1}{c|}{Std} & \multicolumn{1}{c}{Mean} & \multicolumn{1}{c|}{Std} & \multicolumn{1}{c}{Mean} & \multicolumn{1}{c|}{Std} & \multicolumn{1}{c}{Mean} & \multicolumn{1}{c|}{Std} & \multicolumn{1}{c}{Mean} & \multicolumn{1}{c|}{Std} & \multicolumn{1}{c}{Mean} & \multicolumn{1}{c}{Std} \\ \midrule
\bfseries CTD ${(10^{-1})}$ & 0.00e+00 & 0.00e+00 ($\sim$) & 4.08e-01 & 1.49e-03 ($\sim$) & 9.70e+02 & 1.01e+03 ($\sim$) & 4.08e-01 & 1.19e-03 ($+$) & 4.08e-01 & 1.36e-03 ($+$) & \textbf{3.32e-01} & \textbf{3.53e-02} ($\sim$) & \textbf{1.86e-01} & \textbf{3.22e-02} ($\sim$) \\
\bfseries CTD ${(10^{-2})}$ & 1.92e+00 & 1.05e+01 ($\sim$) & 4.09e-01 & 1.45e-03 ($\sim$) & 9.23e+02 & 1.30e+03 ($\sim$) & 4.09e-01 & 1.08e-03 ($\sim$) & 4.09e-01 & 1.53e-03 ($\sim$) & 3.14e-01 & 4.42e-02 ($\sim$) & 1.79e-01 & 3.03e-02 ($\sim$) \\
\bfseries CTD ${(10^{-3})}$ & 1.84e+00 & 6.59e+00 ($\sim$) & \textbf{4.09e-01} & \textbf{1.32e-03} ($\sim$) & \textbf{1.17e+03} & \textbf{1.60e+03} ($\sim$) & 4.10e-01 & 1.49e-03 ($\sim$) & 4.09e-01 & 1.05e-03 ($\sim$) & 3.08e-01 & 5.12e-02 ($\sim$) & 1.81e-01 & 3.07e-02 ($\sim$) \\
\bfseries CTD ${(10^{-5})}$ & 5.13e+00 & 1.36e+01 ($\sim$) & 4.09e-01 & 1.04e-03 ($\sim$) & 9.84e+02 & 1.13e+03 ($\sim$) & 4.10e-01 & 1.43e-03 ($\sim$) & 4.10e-01 & 1.41e-03 ($\sim$) & 3.31e-01 & 3.64e-02 ($\sim$) & 1.83e-01 & 3.09e-02 ($\sim$) \\
\bfseries CTD ${(10^{-6})}$ & 6.05e+00 & 2.74e+01 ($\sim$) & 4.08e-01 & 1.10e-03 ($\sim$) & 1.13e+03 & 1.14e+03 ($\sim$) & 4.09e-01 & 1.46e-03 ($+$) & 4.09e-01 & 1.61e-03 ($+$) & 3.23e-01 & 4.11e-02 ($\sim$) & 1.64e-01 & 1.66e-02 ($\sim$) \\
\bfseries CTD ${(10^{-4})}$ & \textbf{1.31e+01} & \textbf{3.48e+01} & 4.09e-01 & 1.35e-03 & 5.86e+02 & 9.79e+02 & \textbf{4.10e-01} & \textbf{1.24e-03} & \textbf{4.10e-01} & \textbf{1.11e-03} & 3.29e-01 & 3.49e-02 & 1.78e-01 & 3.00e-02 \\
\bottomrule
\toprule
\bfseries Method & \multicolumn{2}{c|}{\bfseries DTLZ1 ($m=3$)} & \multicolumn{2}{c|}{\bfseries DTLZ2 ($m=3$)} & \multicolumn{2}{c|}{\bfseries DTLZ3 ($m=3$)} & \multicolumn{2}{c|}{\bfseries DTLZ4 ($m=3$)} & \multicolumn{2}{c|}{\bfseries DTLZ5 ($m=3$)} & \multicolumn{2}{c|}{\bfseries DTLZ6 ($m=3$)} & \multicolumn{2}{c}{\bfseries DTLZ7 ($m=3$)} \\
 & \multicolumn{1}{c}{Mean} & \multicolumn{1}{c|}{Std} & \multicolumn{1}{c}{Mean} & \multicolumn{1}{c|}{Std} & \multicolumn{1}{c}{Mean} & \multicolumn{1}{c|}{Std} & \multicolumn{1}{c}{Mean} & \multicolumn{1}{c|}{Std} & \multicolumn{1}{c}{Mean} & \multicolumn{1}{c|}{Std} & \multicolumn{1}{c}{Mean} & \multicolumn{1}{c|}{Std} & \multicolumn{1}{c}{Mean} & \multicolumn{1}{c}{Std} \\ \midrule
\bfseries CTD ${(10^{-1})}$ & \textbf{1.28e+04} & \textbf{5.36e+03} ($-$) & 3.55e-01 & 1.37e-01 ($+$) & \textbf{2.71e+06} & \textbf{6.94e+05} ($-$) & 3.64e-01 & 1.29e-01 ($+$) & 1.19e-01 & 5.24e-03 ($+$) & 9.94e-02 & 1.85e-02 ($\sim$) & 2.28e-01 & 8.53e-04 ($\sim$) \\
\bfseries CTD ${(10^{-2})}$ & 1.03e+04 & 6.17e+03 ($-$) & 5.82e-01 & 6.48e-02 ($+$) & 2.67e+06 & 6.73e+05 ($-$) & 5.78e-01 & 7.14e-02 ($+$) & 1.25e-01 & 3.61e-03 ($+$) & 1.06e-01 & 1.38e-02 ($\sim$) & \textbf{2.41e-01} & \textbf{7.72e-02} ($\sim$) \\
\bfseries CTD ${(10^{-3})}$ & 1.13e+04 & 8.90e+03 ($-$) & 6.23e-01 & 6.28e-02 ($\sim$) & 2.20e+06 & 9.50e+05 ($\sim$) & 6.25e-01 & 3.72e-02 ($\sim$) & 1.28e-01 & 1.87e-03 ($\sim$) & 1.07e-01 & 1.45e-02 ($\sim$) & 2.28e-01 & 6.05e-04 ($\sim$) \\
\bfseries CTD ${(10^{-5})}$ & 3.40e+03 & 5.01e+03 ($\sim$) & \textbf{6.56e-01} & \textbf{2.20e-02} ($\sim$) & 1.73e+06 & 8.41e+05 ($\sim$) & 6.37e-01 & 4.38e-02 ($\sim$) & 1.29e-01 & 2.87e-03 ($\sim$) & 1.02e-01 & 1.71e-02 ($\sim$) & 2.28e-01 & 7.83e-04 ($\sim$) \\
\bfseries CTD ${(10^{-6})}$ & 8.67e+03 & 9.52e+03 ($\sim$) & 6.52e-01 & 2.80e-02 ($\sim$) & 1.75e+06 & 8.87e+05 ($\sim$) & \textbf{6.55e-01} & \textbf{2.41e-02} ($\sim$) & \textbf{1.30e-01} & \textbf{2.37e-03} ($\sim$) & 1.02e-01 & 1.62e-02 ($\sim$) & 2.28e-01 & 1.50e-03 ($\sim$) \\
\bfseries CTD ${(10^{-4})}$ & 6.15e+03 & 7.35e+03 & 6.43e-01 & 3.14e-02 & 1.91e+06 & 1.00e+06 & 6.45e-01 & 2.70e-02 & 1.29e-01 & 2.18e-03 & \textbf{1.07e-01} & \textbf{1.57e-02} & 2.28e-01 & 6.42e-04 \\
\bottomrule
\toprule
\bfseries Method & \multicolumn{2}{c|}{\bfseries Four bar truss design} & \multicolumn{2}{c|}{\bfseries Pressure vessel design} & \multicolumn{2}{c|}{\bfseries Hatch cover design} & \multicolumn{2}{c|}{\bfseries Vehicle safety} & \multicolumn{2}{c|}{\bfseries Car side impact} & \multicolumn{2}{c|}{\bfseries LPA} & \multicolumn{2}{c}{\bfseries Summary} \\
 & \multicolumn{1}{c}{Mean} & \multicolumn{1}{c|}{Std} & \multicolumn{1}{c}{Mean} & \multicolumn{1}{c|}{Std} & \multicolumn{1}{c}{Mean} & \multicolumn{1}{c|}{Std} & \multicolumn{1}{c}{Mean} & \multicolumn{1}{c|}{Std} & \multicolumn{1}{c}{Mean} & \multicolumn{1}{c|}{Std} & \multicolumn{1}{c}{Mean} & \multicolumn{1}{c|}{Std} & \multicolumn{2}{c}{$-$/$\sim$/$+$} \\ \midrule
\bfseries CTD ${(10^{-1})}$ & 5.43e+01 & 2.76e-02 ($\sim$) & 1.09e+10 & 1.91e+08 ($\sim$) & 2.18e+04 & 1.31e+01 ($\sim$) & 2.65e+01 & 6.18e-01 ($\sim$) & 3.23e+02 & 1.17e+01 ($\sim$) & 5.77e+07 & 3.53e+06 ($\sim$) & \multicolumn{2}{c}{2/13/5} \\
\bfseries CTD ${(10^{-2})}$ & 5.43e+01 & 3.06e-02 ($\sim$) & 1.10e+10 & 8.35e+07 ($\sim$) & 2.18e+04 & 1.28e+01 ($\sim$) & 2.64e+01 & 5.84e-01 ($\sim$) & \textbf{3.30e+02} & \textbf{8.27e+00} ($\sim$) & 5.81e+07 & 3.39e+06 ($\sim$) & \multicolumn{2}{c}{2/15/3} \\
\bfseries CTD ${(10^{-3})}$ & \textbf{5.43e+01} & \textbf{2.45e-02} ($\sim$) & 1.09e+10 & 1.16e+08 ($\sim$) & 2.18e+04 & 1.19e+01 ($\sim$) & 2.62e+01 & 1.22e+00 ($\sim$) & 2.94e+02 & 1.00e+02 ($\sim$) & 5.79e+07 & 3.78e+06 ($\sim$) & \multicolumn{2}{c}{1/19/0} \\
\bfseries CTD ${(10^{-5})}$ & 5.43e+01 & 2.72e-02 ($\sim$) & 1.09e+10 & 1.95e+08 ($\sim$) & \textbf{2.18e+04} & \textbf{8.30e+00} ($\sim$) & 2.66e+01 & 5.84e-01 ($\sim$) & 3.25e+02 & 1.40e+01 ($\sim$) & 5.77e+07 & 3.25e+06 ($\sim$) & \multicolumn{2}{c}{0/20/0} \\
\bfseries CTD ${(10^{-6})}$ & 5.43e+01 & 5.14e-02 ($+$) & \textbf{1.10e+10} & \textbf{1.40e+08} ($-$) & 2.18e+04 & 1.16e+01 ($\sim$) & 2.58e+01 & 1.65e+00 ($\sim$) & 3.17e+02 & 1.68e+01 ($\sim$) & 5.69e+07 & 4.39e+06 ($\sim$) & \multicolumn{2}{c}{1/16/3} \\
\bfseries CTD ${(10^{-4})}$ & 5.43e+01 & 2.80e-02 & 1.09e+10 & 1.91e+08 & 2.18e+04 & 1.52e+01 & \textbf{2.66e+01} & \textbf{6.20e-01} & 3.26e+02 & 1.07e+01 & \textbf{5.82e+07} & \textbf{3.97e+06} & \multicolumn{2}{c}{--/--/--} \\
\bottomrule
\end{tabular}
}
\end{table*}

\subsection{Using an Alternative Acquisition Function} \label{appendix:sec:acf}

\begin{table*}[!ht]
\centering
\caption{The HV results (mean and standard deviation) of the eight methods on the 20 benchmark and
real-world problems under noisy settings. 
The method with the best mean HV is highlighted in bold. The symbols ``$-$'', ``$\sim$'' and ``$+$'' indicate that CTD$_{tch}$ is statistically better than, equivalent to, and worse than EHVI-based CTD, respectively.}
\label{tab:noisy_100_acf}
\resizebox{\textwidth}{!}{
\begin{tabular}{l | ll| ll| ll| ll| ll| ll|ll}
\toprule
\bfseries Method & \multicolumn{2}{c|}{\bfseries DTLZ1 ($m=2$)} & \multicolumn{2}{c|}{\bfseries DTLZ2 ($m=2$)} & \multicolumn{2}{c|}{\bfseries DTLZ3 ($m=2$)} & \multicolumn{2}{c|}{\bfseries DTLZ4 ($m=2$)} & \multicolumn{2}{c|}{\bfseries DTLZ5 ($m=2$)} & \multicolumn{2}{c|}{\bfseries DTLZ6 ($m=2$)} & \multicolumn{2}{c}{\bfseries DTLZ7 ($m=2$)} \\
 & \multicolumn{1}{c}{Mean} & \multicolumn{1}{c|}{Std} & \multicolumn{1}{c}{Mean} & \multicolumn{1}{c|}{Std} & \multicolumn{1}{c}{Mean} & \multicolumn{1}{c|}{Std} & \multicolumn{1}{c}{Mean} & \multicolumn{1}{c|}{Std} & \multicolumn{1}{c}{Mean} & \multicolumn{1}{c|}{Std} & \multicolumn{1}{c}{Mean} & \multicolumn{1}{c|}{Std} & \multicolumn{1}{c}{Mean} & \multicolumn{1}{c}{Std} \\ \midrule
\bfseries CTD$_{tch}$ & 0.00e+00 & 0.00e+00 ($\sim$) & 6.38e-02 & 5.88e-02 ($\sim$) & \textbf{4.43e+02} & \textbf{8.41e+02} ($\sim$) & 4.19e-02 & 4.82e-02 ($\sim$) & 4.88e-02 & 5.12e-02 ($\sim$) & 1.02e-01 & 6.21e-02 ($\sim$) & \textbf{5.90e-02} & \textbf{1.15e-01} ($-$) \\
\bfseries CTD & \textbf{8.72e-03} & \textbf{4.78e-02} & \textbf{7.59e-02} & \textbf{5.95e-02} & 3.69e+02 & 6.54e+02 & \textbf{5.68e-02} & \textbf{4.74e-02} & \textbf{6.70e-02} & \textbf{4.69e-02} & \textbf{1.33e-01} & \textbf{7.48e-02} & 1.48e-02 & 5.13e-02 \\
\bottomrule
\toprule
\bfseries Method & \multicolumn{2}{c|}{\bfseries DTLZ1 ($m=3$)} & \multicolumn{2}{c|}{\bfseries DTLZ2 ($m=3$)} & \multicolumn{2}{c|}{\bfseries DTLZ3 ($m=3$)} & \multicolumn{2}{c|}{\bfseries DTLZ4 ($m=3$)} & \multicolumn{2}{c|}{\bfseries DTLZ5 ($m=3$)} & \multicolumn{2}{c|}{\bfseries DTLZ6 ($m=3$)} & \multicolumn{2}{c}{\bfseries DTLZ7 ($m=3$)} \\
 & \multicolumn{1}{c}{Mean} & \multicolumn{1}{c|}{Std} & \multicolumn{1}{c}{Mean} & \multicolumn{1}{c|}{Std} & \multicolumn{1}{c}{Mean} & \multicolumn{1}{c|}{Std} & \multicolumn{1}{c}{Mean} & \multicolumn{1}{c|}{Std} & \multicolumn{1}{c}{Mean} & \multicolumn{1}{c|}{Std} & \multicolumn{1}{c}{Mean} & \multicolumn{1}{c|}{Std} & \multicolumn{1}{c}{Mean} & \multicolumn{1}{c}{Std} \\ \midrule
\bfseries CTD$_{tch}$ & 6.61e+03 & 6.16e+03 ($\sim$) & 5.20e-02 & 5.54e-02 ($+$) & \textbf{1.68e+06} & \textbf{6.09e+05} ($\sim$) & 5.40e-02 & 5.44e-02 ($+$) & 5.21e-03 & 8.78e-03 ($+$) & 2.18e-02 & 2.95e-02 ($+$) & 0.00e+00 & 0.00e+00 ($\sim$) \\
\bfseries CTD & \textbf{8.33e+03} & \textbf{6.55e+03} & \textbf{9.15e-02} & \textbf{4.19e-02} & 1.43e+06 & 6.62e+05 & \textbf{8.17e-02} & \textbf{4.46e-02} & \textbf{1.47e-02} & \textbf{1.20e-02} & \textbf{5.16e-02} & \textbf{3.09e-02} & \textbf{0.00e+00} & \textbf{0.00e+00} \\
\bottomrule
\toprule
\bfseries Method & \multicolumn{2}{c|}{\bfseries Four bar truss design} & \multicolumn{2}{c|}{\bfseries Pressure vessel design} & \multicolumn{2}{c|}{\bfseries Hatch cover design} & \multicolumn{2}{c|}{\bfseries Vehicle safety} & \multicolumn{2}{c|}{\bfseries Car side impact} & \multicolumn{2}{c|}{\bfseries LPA} & \multicolumn{2}{c}{\bfseries Summary} \\
 & \multicolumn{1}{c}{Mean} & \multicolumn{1}{c|}{Std} & \multicolumn{1}{c}{Mean} & \multicolumn{1}{c|}{Std} & \multicolumn{1}{c}{Mean} & \multicolumn{1}{c|}{Std} & \multicolumn{1}{c}{Mean} & \multicolumn{1}{c|}{Std} & \multicolumn{1}{c}{Mean} & \multicolumn{1}{c|}{Std} & \multicolumn{1}{c}{Mean} & \multicolumn{1}{c|}{Std} & \multicolumn{2}{c}{$-$/$\sim$/$+$} \\ \midrule
\bfseries CTD$_{tch}$ & \textbf{5.19e+01} & \textbf{9.48e-01} ($-$) & 1.05e+10 & 5.25e+08 ($\sim$) & \textbf{2.15e+04} & \textbf{2.54e+02} ($\sim$) & \textbf{2.64e+01} & \textbf{5.92e-01} ($-$) & \textbf{2.70e+02} & \textbf{1.67e+01} ($-$) & 4.11e+07 & 4.67e+06 ($+$) & \multicolumn{2}{c}{4/11/5} \\
\bfseries CTD & 5.11e+01 & 1.08e+00 & \textbf{1.06e+10} & \textbf{2.83e+08} & 2.15e+04 & 1.70e+02 & 2.57e+01 & 6.86e-01 & 2.57e+02 & 1.95e+01 & \textbf{4.52e+07} & \textbf{5.62e+06} & \multicolumn{2}{c}{--/--/--} \\
\bottomrule
\end{tabular}
}
\end{table*}

\begin{table*}[!ht]
\centering
\caption{The HV results (mean and standard deviation) of the two methods on 14 problems with $d=50$. 
The method with the best mean HV is highlighted in bold. The symbols ``$-$'', ``$\sim$'' and ``$+$'' indicate that the method is statistically better than, equivalent to, and worse than CTD, respectively.}
\label{tab:hd_acf}
\resizebox{\textwidth}{!}{
\begin{tabular}{l | ll| ll| ll| ll| ll| ll| ll}
\toprule
\bfseries Method & \multicolumn{2}{c|}{\bfseries DTLZ1 ($m=2$)} & \multicolumn{2}{c|}{\bfseries DTLZ2 ($m=2$)} & \multicolumn{2}{c|}{\bfseries DTLZ3 ($m=2$)} & \multicolumn{2}{c|}{\bfseries DTLZ4 ($m=2$)} & \multicolumn{2}{c|}{\bfseries DTLZ5 ($m=2$)} & \multicolumn{2}{c|}{\bfseries DTLZ6 ($m=2$)} & \multicolumn{2}{c}{\bfseries DTLZ7 ($m=2$)} \\
 & \multicolumn{1}{c}{Mean} & \multicolumn{1}{c|}{Std} & \multicolumn{1}{c}{Mean} & \multicolumn{1}{c|}{Std} & \multicolumn{1}{c}{Mean} & \multicolumn{1}{c|}{Std} & \multicolumn{1}{c}{Mean} & \multicolumn{1}{c|}{Std} & \multicolumn{1}{c}{Mean} & \multicolumn{1}{c|}{Std} & \multicolumn{1}{c}{Mean} & \multicolumn{1}{c|}{Std} & \multicolumn{1}{c}{Mean} & \multicolumn{1}{c}{Std} \\ \midrule
\bfseries CTD$_{tch}$ & 1.53e+05 & 1.97e+04 ($+$) & \textbf{3.16e-01} & \textbf{1.68e-01} ($-$) & 2.71e+05 & 7.69e+04 ($\sim$) & \textbf{1.69e-01} & \textbf{1.09e-01} ($\sim$) & \textbf{1.12e+00} & \textbf{2.39e-01} ($-$) & \textbf{1.64e-01} & \textbf{1.00e-01} ($\sim$) & \textbf{3.63e-01} & \textbf{1.83e-01} ($-$)\\
\bfseries CTD & \textbf{1.94e+05} & \textbf{2.83e+04} & 1.84e-01 & 1.59e-01 & \textbf{2.97e+05} & \textbf{7.55e+04} & 1.29e-01 & 1.17e-01 & 7.42e-01 & 4.11e-01 & 1.56e-01 & 9.46e-02 & 2.11e-01 & 2.45e-02\\
\bottomrule
\toprule
\bfseries Method & \multicolumn{2}{c|}{\bfseries DTLZ1 ($m=3$)} & \multicolumn{2}{c|}{\bfseries DTLZ2 ($m=3$)} & \multicolumn{2}{c|}{\bfseries DTLZ3 ($m=3$)} & \multicolumn{2}{c|}{\bfseries DTLZ4 ($m=3$)} & \multicolumn{2}{c|}{\bfseries DTLZ5 ($m=3$)} & \multicolumn{2}{c|}{\bfseries DTLZ6 ($m=3$)} & \multicolumn{2}{c}{\bfseries DTLZ7 ($m=3$)} \\
 & \multicolumn{1}{c}{Mean} & \multicolumn{1}{c|}{Std} & \multicolumn{1}{c}{Mean} & \multicolumn{1}{c|}{Std} & \multicolumn{1}{c}{Mean} & \multicolumn{1}{c|}{Std} & \multicolumn{1}{c}{Mean} & \multicolumn{1}{c|}{Std} & \multicolumn{1}{c}{Mean} & \multicolumn{1}{c|}{Std} & \multicolumn{1}{c}{Mean} & \multicolumn{1}{c|}{Std} & \multicolumn{1}{c}{Mean} & \multicolumn{1}{c}{Std} \\ \midrule
\bfseries CTD$_{tch}$ & 1.10e+08 & 1.95e+07 ($+$) & \textbf{1.72e+01} & \textbf{2.47e+00} ($-$) & 4.61e+08 & 1.09e+08 ($\sim$) & \textbf{7.68e+00} & \textbf{1.61e+00} ($-$) & \textbf{1.72e+01} & \textbf{2.01e+00} ($-$) & 1.64e-02 & 2.91e-02 ($+$) & \textbf{6.34e-01} & \textbf{2.80e-01} ($-$)\\
\bfseries CTD & \textbf{1.80e+08} & \textbf{1.11e+07} & 1.44e+01 & 2.97e+00 & \textbf{5.09e+08} & \textbf{9.43e+07} & 6.47e+00 & 2.16e+00 & 1.49e+01 & 4.24e+00 & \textbf{6.41e-02} & \textbf{1.56e-02} & 2.50e-01 & 6.98e-02\\
\bottomrule
\end{tabular}
}
\end{table*}

\begin{table*}[!ht]\tiny
\centering
\caption{The summary of statistical results for the high-dimensional settings in Table~\ref{tab:hd_acf}. Here, the left, median, and right numbers are the counts of test problems where the CTD was statistically worse, equivalent, or better to CTD$_{tch}$, respectively.}
\label{tab:hd_summary_acf}
\begin{tabular}{lc}
\toprule
 & \bfseries CTD$_{tch}$ \\
\midrule
\bfseries CTD & 7/4/3 \\
\bottomrule
\end{tabular}
\end{table*}

\section{Extensions}\label{appendix:sec:extensions}

The preceding discussion has addressed multi-objective optimisation problems in which evaluations are performed under noiseless, noisy, and high-dimensional settings. 
However, not all multi-objective settings conform to these scenarios. 
To accommodate a broader class of problems, we propose several extensions that enable the methodology to handle more optimisation scenarios. 



\paragraph{Constrained Bayesian Optimisation (CBO).}

Existing CBO methods can be divided into explicit and implicit methods~\citep{amini2024constrained}. 
Explicit methods first estimate the feasible region from constraint surrogate models and then optimise a standard acquisition function within this region~\citep{sasena2002exploration,xu2023constrained}. 
In contrast, implicit methods avoid explicitly solving a constrained problem by incorporating feasibility into the acquisition function. 
For example, they may multiply improvement by the probability of feasibility~\citep{gelbart2014bayesian,letham2019constrained}, or use penalty~\citep{ariafar2019admmbo,pourmohamad2022bayesian,amini2023bayesian,picheny2016bayesian}.
Our CTD can be naturally extended to the constrained setting through the above techniques.

\paragraph{Multi-Fidelity Bayesian Optimisation (MFBO).} 

In many real-world optimisation scenarios, the evaluation is often available at multiple fidelity levels, where increasing fidelity typically leads to improved accuracy at the expense of higher computational cost. 
Many MFBO methods have been proposed to tackle such optimisation problems~\cite{belakaria2020multi,kandasamy2017multi,li2020multi,moss2021gibbon,song2019general,takeno2020multi,wu2020practical,zhang2017information}. 
Our proposed CTD can be potentially extended to the multi-fidelity setting by integrating prior techniques, e.g., building multiple surrogate models of different levels of fidelity. 

\end{document}

%% file: math_commands.tex
\usepackage{amsmath,amsfonts,bm}

\def\eqref#1{equation~\ref{#1}}

\def\1{\bm{1}}

\DeclareMathAlphabet{\mathsfit}{\encodingdefault}{\sfdefault}{m}{sl}
\SetMathAlphabet{\mathsfit}{bold}{\encodingdefault}{\sfdefault}{bx}{n}

